\documentclass[10pt,twocolumn,letterpaper]{article}

\usepackage{cvpr}              

\usepackage{colortbl} 
\usepackage{soul} 

\definecolor{cvprblue}{rgb}{0.21,0.49,0.74}
\usepackage[pagebackref,breaklinks,colorlinks,allcolors=cvprblue]{hyperref}

\usepackage[flushmargin]{footmisc}
\usepackage{endnotes}

\usepackage{mathrsfs} 
\usepackage{booktabs}
\usepackage{kotex} 
\usepackage{lipsum} 
\usepackage{duckuments} 
\def\paperID{15984} 
\def\confName{CVPR}
\def\confYear{2025}
\newcommand{\sname}{BootComp} 

\newcommand\blfootnote[1]{%
  \begingroup
  \renewcommand\thefootnote{}\footnote{#1}%
  \addtocounter{footnote}{-1}%
  \endgroup
}

\newcommand*{\ShowNotes}{} 

\ifdefined\ShowNotes
  \newcommand{\colornote}[3]{{\color{#1}\bf{#2: #3}\normalfont}}
\else
  \newcommand{\colornote}[3]{}
\fi

\newcommand{\rvx}{\mathbf{x}}
\newcommand{\rvy}{\mathbf{y}}

\newcommand{\rvc}{\mathbf{c}}

\newcommand{\rvm}{\mathbf{m}}

\title{Controllable Human Image Generation with Personalized Multi-Garments}

\author{
Yisol Choi$^{1}$ \hspace{10pt} Sangkyung Kwak$^{1,3}$ \hspace{10pt} Sihyun Yu$^{1}$ \hspace{10pt} Hyungwon Choi$^{2}$ \hspace{10pt} Jinwoo Shin$^{1}$ \\
$^{1}$KAIST \hspace{10pt} $^{2}$OMNIOUS.AI  \hspace{10pt} $^{3}$Scaled Foundations \\
{\tt\small \{yisol.choi, skkwak9806, sihyun.yu, jinwoos\}@kaist.ac.kr, hyungwon.choi@omnious.com}
}

\begin{document}
\twocolumn[{%
\maketitle
\vspace{-0.05in}
\includegraphics[width=.955\textwidth]{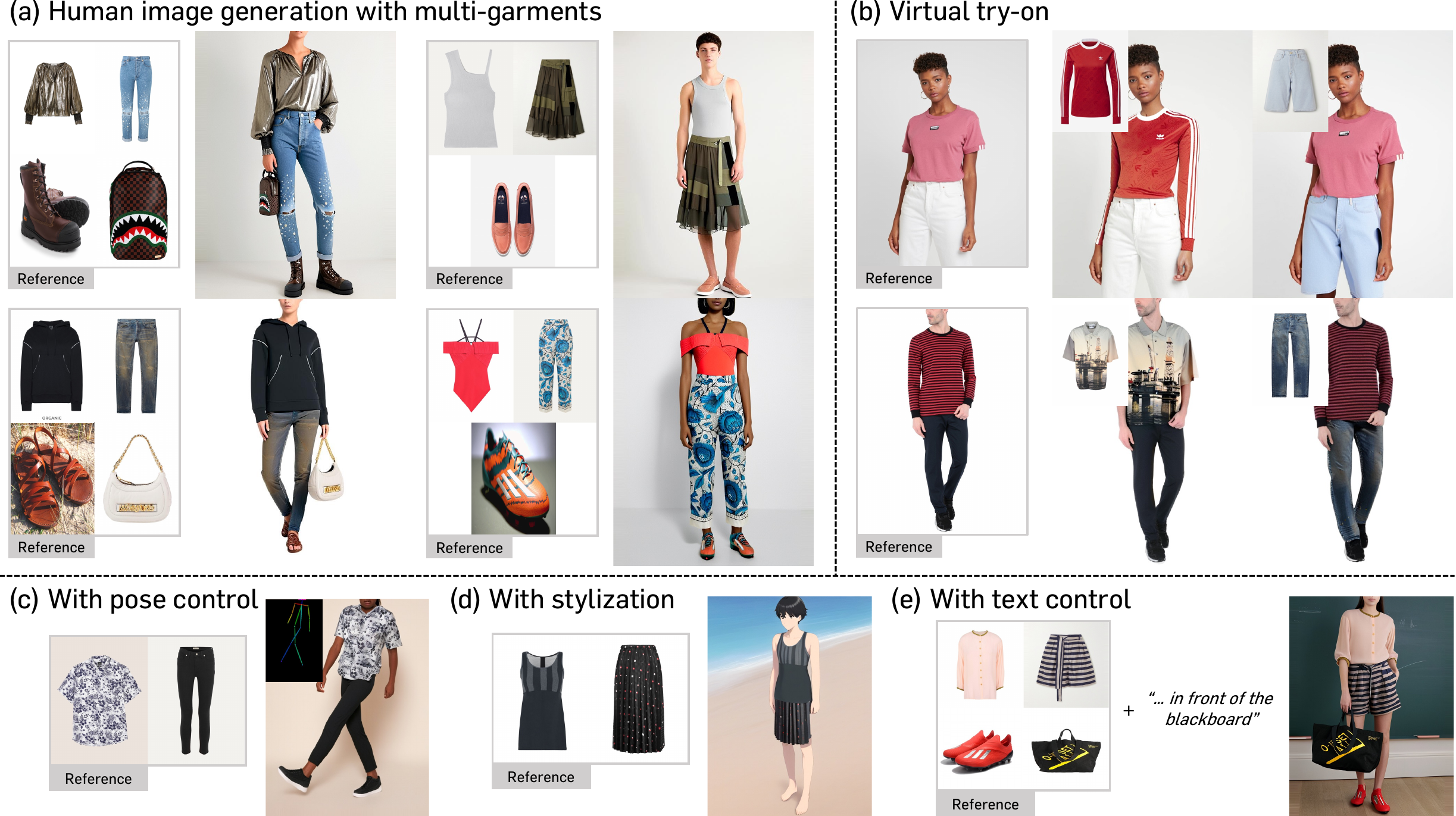}    
\captionof{figure}{\textbf{Generated images by \sname.} (a) \sname~generates high-quality human images wearing multiple reference garments, with support for extended categories such as bag, shoes, even in unusual garment combinations (\emph{e.g.}, swimming suit with soccer cleats). We show \sname's~generalization capability through various conditional image generations, such as (b) virtual try-on, (c) pose guided generation, (d) stylization, and (e) text guided generation, even though \sname~is not directly trained or fine-tuned for each task. }

\vspace{0.2in}
\label{fig:teaser}
}]

\begin{figure*}[!t]
    \small\centering
    \includegraphics[width=\textwidth]{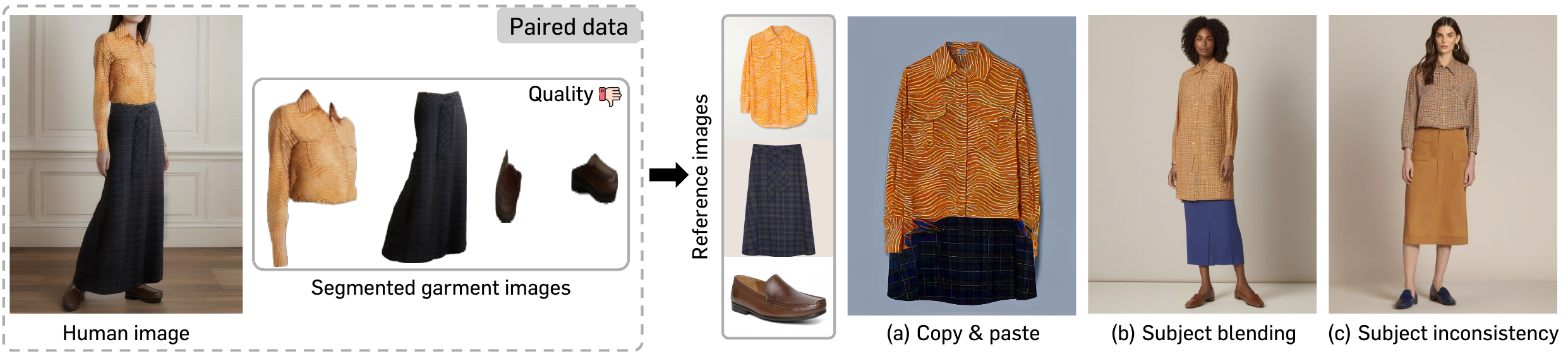}
    \vspace{-13pt}
    \caption{\textbf{Limitations of previous data curation approaches used in controllable generation.} 
    Previous approaches on controllable generation often use a paired dataset consisting of low-quality segmented garments and human images for training. It leads to several undesirable artifacts as shown in right (generated with baselines). 
    For example, garments are directly replicated from the reference images in (a), shirts and skirts are blended together in (b), and generated skirts fail to resemble the reference in (c).
    }

    \label{fig:limitations}
    \vspace{-13.5pt}
\end{figure*}

\begin{abstract}
\blfootnote{Project page: \href{https://omnious.github.io/BootComp}{https://omnious.github.io/BootComp}}
We present \sname, a novel framework based on text-to-image diffusion models for controllable human image generation with multiple reference garments.
Here, the main bottleneck is data acquisition for training: collecting a large-scale dataset of high-quality reference garment images per human subject is quite challenging, i.e., ideally, one needs to manually gather every single garment photograph worn by each human.
To address this, we propose a data generation pipeline to construct a large synthetic dataset, consisting of human and multiple-garment pairs, by introducing a model to extract any reference garment images from each human image.
To ensure data quality, we also propose a filtering strategy to remove undesirable generated data based on measuring perceptual similarities between the garment presented in human image and extracted garment.
Finally, by utilizing the constructed synthetic dataset, we train a diffusion model having two parallel denoising paths that use multiple garment images as conditions to generate human images while preserving their fine-grained details.
We further show the wide-applicability of our framework by adapting it to 
different types of reference-based generation in the fashion domain, including virtual try-on, and controllable human image generation with other conditions, e.g., pose, face, etc.
\end{abstract}
    
\vspace{-0.5in}
\section{Introduction}
\label{sec:intro}

Recent advances in text-to-image (T2I) diffusion models~\citep{rombach2022high, podell2023sdxl, esser2024scaling} have shown great progress in numerous challenging real-world scenarios, such as personalized generation~\citep{ruiz2022dreambooth, lee2024direct}, style transfer~\citep{hertz2024style, wang2023stylediffusion}, image editing~\citep{brooks2023instructpix2pix, meng2021sdedit, hertz2022prompt}, and compositional image generation~\citep{kosmos-g,wang2024ms,nie2024BlobGEN}. These remarkable successes have provided great potential to aid users in a variety of creative pursuits~\citep{ko2023large}.

Among them, \emph{controllable human image generation} using T2I diffusion models~\citep{huang2024parts2whole} can provide lots of intriguing use cases in real-world scenarios. Specifically, by training a model capable of creating human images conditioned on a variety of garments,
one can enjoy diverse applications such as outfit recommendations for users, generating fashion models for clothing brands, or virtual try-on~\citep{morelli2023ladi,kim2024stableviton,choi2024improving}, through a \emph{single unified framework}.

One can consider fine-tuning T2I models and image encoders~\citep{radford2021learning, oquab2024dinov} using curated paired image datasets that consist of condition garments and the target human images~\citep{huang2024parts2whole}. However, hand-collecting multiple garment photographs worn by human is labor-intensive. 
Prior works~\cite{kosmos-g, huang2024resolving,ye2023ip-adapter} have attempted to obtain the pair images by extracting all reference objects from real images, segmenting out each object from the original images.
However, this data curation protocol makes curated garments have exactly the same shape with their appearance in the target human image. Thus, generated images are likely to suffer from copy-and-paste problem: they easily generate exactly the same image in generated samples without altering pose or appearance (see (a) in \cref{fig:limitations}).
To mitigate this issue, several works propose to curate data from videos by doing segmentation from different video frames that contain the same objects~\citep{wang2024ms,chen2023anydoor}. However, collecting such paired datasets in large amounts is challenging and often results in low-quality reference images; thereby, the trained model fails to generalize and suffers from subject blending or inconsistency within the images~\citep{wang2024ms} (see (b), (c) in \cref{fig:limitations}). Such drawbacks become more critical in practical scenarios related to human image generation, as the model must generate human images with diverse poses while accurately preserving the details of each garment.

\begin{figure*}[!t]
    \small\centering
    \includegraphics[width=\textwidth]{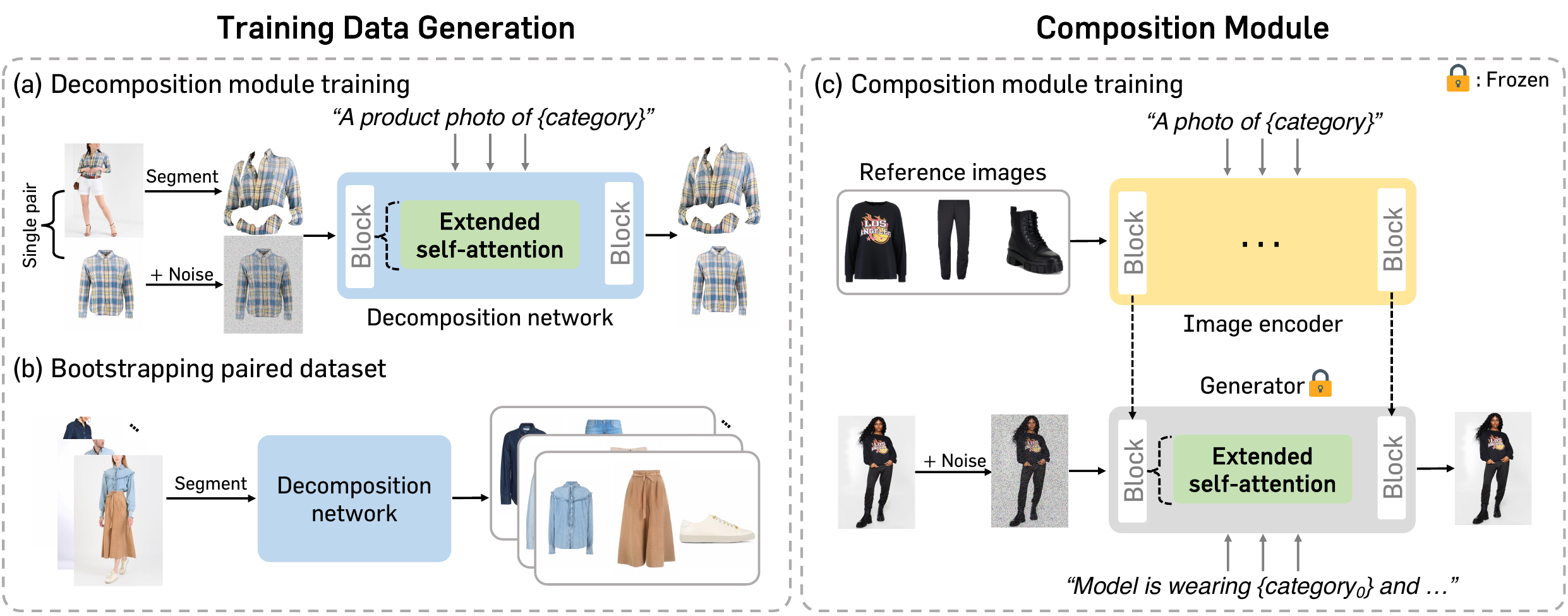}
    \vspace{-0.17in}
    \caption{\textbf{Overview of \sname.} 
    We propose a two-stage framework: synthetic data generation and composition module training for controllable human image generation.   
    (a) We train a decomposition network that maps from a segmented garment image to a product garment image. (b) We bootstrap synthetic paired data of human and multiple garment images. (c) We finally train our composition module with the synthetic paired dataset enabling it to generate human images with multiple reference garment images.}
    \label{fig:overview}
    \vspace{-0.12in}
\end{figure*}

\vspace{0.02in}
\noindent
\textbf{Contribution.}
We address the aforementioned shortcomings by presenting \textit{Bootstrapping paired data for Compositional and controllable human image generation}~(\sname), a novel framework for controllable human image generation using T2I diffusion models. Specifically, it is a two-stage framework (see \cref{fig:overview} for illustration):
\vspace{0.02in}
\begin{itemize}[topsep=0.0pt,itemsep=1.2pt,leftmargin=2.5mm]
    \item \emph{Synthetic data generation}: We first propose a high-quality synthetic data generation pipeline for training controllable human image generation model. We achieve this by introducing a decomposition module, which is a mapping from a single garment worn by a human to a product view of the garment image. We train this model with a paired dataset of \textit{single} reference garment and human image (\eg, shirts and human wearing those shirts), which is easy to collect~\cite{choi2021viton,morelli2022dress,lepage2023lrvsf}. Using this model, we bootstrap synthetic paired data at scale from a large number of human images; thus, each pair consists of a human image and all garment images that the human is wearing. To ensure high-quality data, we also present a filtering strategy that further improves the data quality based on measuring the perceptual similarities between the original segmentation results and the data generated from the decomposition module.
    \item \emph{Composition module}: We also present a fine-tuning scheme of T2I diffusion models for our goal using the synthetic dataset. We use two T2I diffusion models: one serves as an image encoder to extract garment features and the other one functions as a generator to create human images. We only train the encoder model, employing an extended self-attention mechanism to generator for conditioning garment images. Since we keep the generator frozen during the training, \sname~can be attached to various adapter modules or replaced with pre-trained models specialized to generate images with different styles. This enables \sname~to provide various applications (\eg, pose-guided or cartoon-style generation) for free without requiring any additional fine-tuning. 
\end{itemize}

We demonstrate the effectiveness of \sname~in terms of garment fidelity and compositionality through extensive experiments. For example, \sname~shows 30\% improvement on MP-LPIPS~\citep{chen2024magic} than the previous state-of-the-art methods.
Moreover, our \sname~is extensively applied to various conditional human image generations in the fashion domain, such as virtual try-on and controllable human image generation with other conditions, such as faces and poses. We also highlight the generalization capabilities of \sname~across different image domains, generating human images in various styles like cartoons.

\section{Background} 
\subsection{Diffusion Models} \label{DMs} 
Diffusion models~\cite{ho2020denoising, sohl2015deep, karras2022elucidating, karras2023analyzing} are a type of generative model consisting of a forward process and a reverse process. Specifically, diffusion models learn the reverse process of the forward process, where the forward process is defined as a Markov chain that gradually adds Gaussian noise to data. Starting from Gaussian noise, The sampling is done with a learned reverse process of this forward process.

Formally, let $\mathbf{x}_0$ represent a data instance (\emph{e.g.}, an image or a latent vector from an autoencoder’s output~\cite{rombach2022high}). Diffusion models consider a pre-defined forward process $q(\rvx_{t} | \rvx_{0})$ given a closed form as a normal distribution $\mathcal{N}(\alpha_t\mathbf{x}_0, \sigma_t^2\mathbf{I})$, so the sampling can be done from Gaussian distribution $\boldsymbol{\epsilon} \sim \mathcal{N}(\boldsymbol{0}, \mathbf{I})$ using reparametrization to have $\mathbf{x}_t = \alpha_t\mathbf{x}_0 + \sigma_t \boldsymbol{\epsilon}$. Here, $\{\alpha_t\}_{t=1}^T$ and $\{\sigma_t\}_{t=1}^T$ are pre-defined decreasing and increasing noise scheduling sequences (respectively) for $t=1,\ldots, T$ that let $p(\mathbf{x}_T)$ converge a distribution close to Gaussian distribution $\mathcal{N}(\boldsymbol{0}, \mathbf{I})$. 

Learning the reverse process $p_{\theta}(\rvx_{t-1} |  \rvx_t)$ of a diffusion model is equivalent to learning a score function of perturbed data distribution (through score matching~\cite{hyvarinen2005estimation}), typically achieved via an $\epsilon$-noise prediction loss~\cite{ho2020denoising} by training a denoising autoencoder. 
Specifically, one can formulate the training objective of the diffusion model as:
\begin{align*}
    \mathcal{L}_{\textrm{DM}} = \mathbb{E}_{ \boldsymbol{\epsilon}\sim\mathcal{N}(\boldsymbol{0},\mathbf{I}), \,t\sim\mathcal{U}[0,T]}\big[\,\omega(t) \|\boldsymbol{\epsilon}_\theta(\mathbf{x}_t;t) - \boldsymbol{\epsilon}\|_2^2 \,\big]\text{,}
\end{align*}
where $\omega(t) >0 $ is a weight function at each timestep $t$ and $\mathcal{U}[0,T]$ denotes a uniform distribution.

After training, data sampling can be done using the learned reverse process. Specifically, starting from $\mathbf{x}_T\sim \mathcal{N}(\boldsymbol{0},\sigma_T^2\mathbf{I})$, the model gradually denoises $\mathbf{x}_{t}$ to $\mathbf{x}_{t-1}$ for each $t$, until $\mathbf{x}_0$ is drawn from the data distribution.

\subsection{Text-to-Image (T2I) Diffusion Models} \label{T2I}
Text-to-image (T2I) diffusion models~\cite{saharia2022photorealistic, rombach2022high, esser2024scaling} are text-conditional diffusion models $\boldsymbol{\epsilon}_\theta(\mathbf{x}_t;\rvc, t) $ that generate an image $\rvx_0$ conditioned on a given text prompt $\rvc$. This prompt is usually provided as a text representation encoded by pre-trained text encoders, such as T5~\cite{raffel2020exploring} or CLIP~\cite{radford2021learning}. Commonly, T2I diffusion models employ convolutional U-Net architectures combined with attention layers~\cite{ho2020denoising, song2020score} to condition the model on texts.
Among T2I diffusion models, Stable Diffusion~\citep[SD;][]{rombach2022high} is one of the de-facto T2I diffusion models that generates high-quality images. We mainly use Stable Diffusion XL (SDXL)~\citep{podell2023sdxl}, one of the SD variants. However, our framework is model-agnostic and can be adapted to any other T2I diffusion models.

\section{Method}

Let $\mathbf{X}=\{\rvx_1, \ldots ,\rvx_N\}$ be a set of $N \gg 1$ reference garment images (\eg, shirt, pants, \etc)
and $\rvy$ be a human image that is wearing $\rvx_1, \ldots ,\rvx_N$. Our goal is to learn a conditional distribution $p(\rvy | \mathbf{X})$---we train a conditional generative model $g_{\theta}(\mathbf{X})=\rvy$ that generates human image $\rvy$ wearing arbitrary garment images $\mathbf{X}$ given as a condition.

One straightforward direction is to train the model $g_{\theta}$ using a paired dataset $\mathcal{D} = \{(\mathbf{X}^{i}, \rvy^{i})\}_{i=1}^{d}$ with a dataset size $d>0$, where each $\mathbf{X}^{i}=\{\rvx_{1}^{i},\dots,\rvx_{N_i}^{i}\}$ consists of $N_i$ different number of reference images. However, this approach suffer from data acquisition problem: collecting all of the reference garment images of a given human image is wearing is challenging. In practice, there usually exists a single reference image, \ie, $N_i$ mostly becomes 1 (\eg, a human and pants that he/she is wearing). Thus, the model trained with this data easily lacks compositional generalization capability at inference time, \ie, the trained model $g_{\theta}$ fails to generate the human image with large number of garments.

To tackle this data curation problem, we introduce an additional decomposition network $f_{\phi}$ that can extract reference images from a given human image. By doing so, we generate a synthetic dataset $\tilde{\mathcal{D}}$, where each $(\tilde{\mathbf{X}}^{i}, \rvy^{i}) \in \tilde{\mathcal{D}}$ satisfies $|\tilde{\mathbf{X}}^i| \gg 1$ and $\rvy^{i}$ is in the original dataset. We then train the conditional generative model $g_{\theta}$ using this synthetic dataset.
Here, we also introduce a filtering strategy to improve the quality of the synthetic dataset $\tilde{\mathcal{D}}$ generated from  $f_{\phi}$, by removing low-quality extraction results.

In the rest of this section, we explain our~\sname~in detail. In Section~\ref{sec:datagen}, we describe the training data generation process, introducing our decomposition network $f_{\phi}$, which is used for synthetic data generation, and explaining our data filtering strategy. Finally, in Section~\ref{sec:comp}, we explain the details of our network for our original goal of controllable generation trained with the synthetic dataset.

\subsection{Training data generation} \label{sec:datagen}

\noindent\textbf{Decomposition module. }
Our decomposition module generates a \emph{single} garment image in a product view, denoted as $\rvx$, from a garment of category $\rvm$ that human $\rvy$ is wearing. We consider this mapping as an image-to-image translation problem: generating the reference garment image $\rvx$ from the portion of person image $\rvy$ that falls into category $\rvm$.

To achieve this, we initialize a diffusion model $f_{\phi}$ as a pre-trained text-to-image diffusion model and fine-tune it with the following objective:
\begin{align}
    \mathcal{L}(\phi):= \mathbb{E} \Big[ \omega(t) \big|\big| f_\phi(\mathbf{x}_t;\mathbf{c},t, \rvx^s) -
    \boldsymbol{\epsilon} \big|\big|_2^2 \Big],
\end{align}
where $\rvx^s = S(\rvy,\rvm)$ is a segmented garment part using an off-the-shelf human parsing model $S$ \citep{xie2021segformer}, and we let a text prompt $\rvc$ be ``A product photo of \textit{\{category\}}'' to extensively leverage the prior knowledge of the T2I diffusion model.

\begin{figure}[t]
\includegraphics[width=\linewidth]
{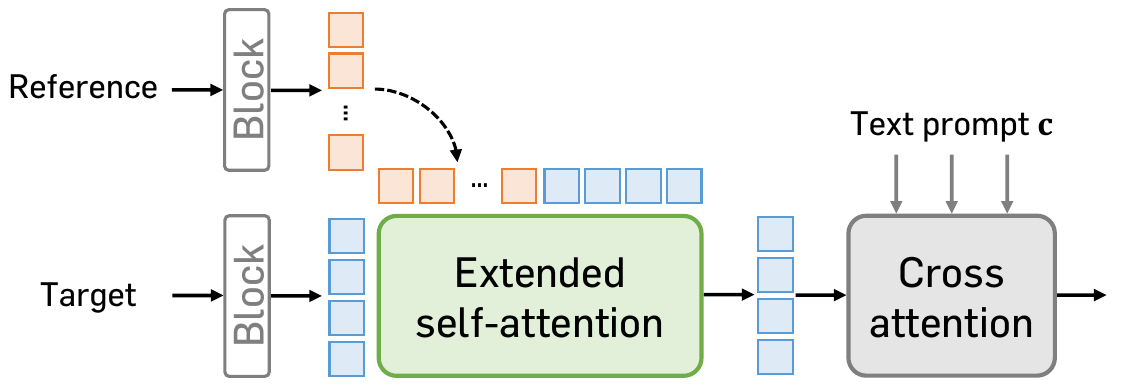}
\captionof{figure}{\textbf{Extended self-attention architecture.} In a extended self-attention layer, \sethlcolor{Orange!15} \hl{reference hidden states} are concatenated with the \sethlcolor{RoyalBlue!15} \hl{target hidden states} in the key and value matrices. This architecture enables injecting reference image features within the target image. Note that decomposition module also uses same structure but works within a single network.}
\label{fig:attn_arch}
\end{figure}

To condition the model on an image $\rvx^s$, we utilize the pretrained diffusion model as an image encoder, which can extract rich features and can preserve the fine-details (\eg, small logos). Specifically, for each self-attention layer in the model, we concatenate the corresponding key and value vectors computed with $\rvx^s$, so the self-attention in the forwarding path of $\mathbf{x}_t$ can be conditioned on $\rvx^s$ (see Fig. \ref{fig:attn_arch}).

Finally, note that training $f_{\phi}$ can be done with the dataset $\mathcal{D}$ which consists of a pair of \emph{single} reference garment and a human image, because we train the model to extract a \emph{single} reference garment from the human image.

\begin{figure}[!bht]
\includegraphics[width=\linewidth]{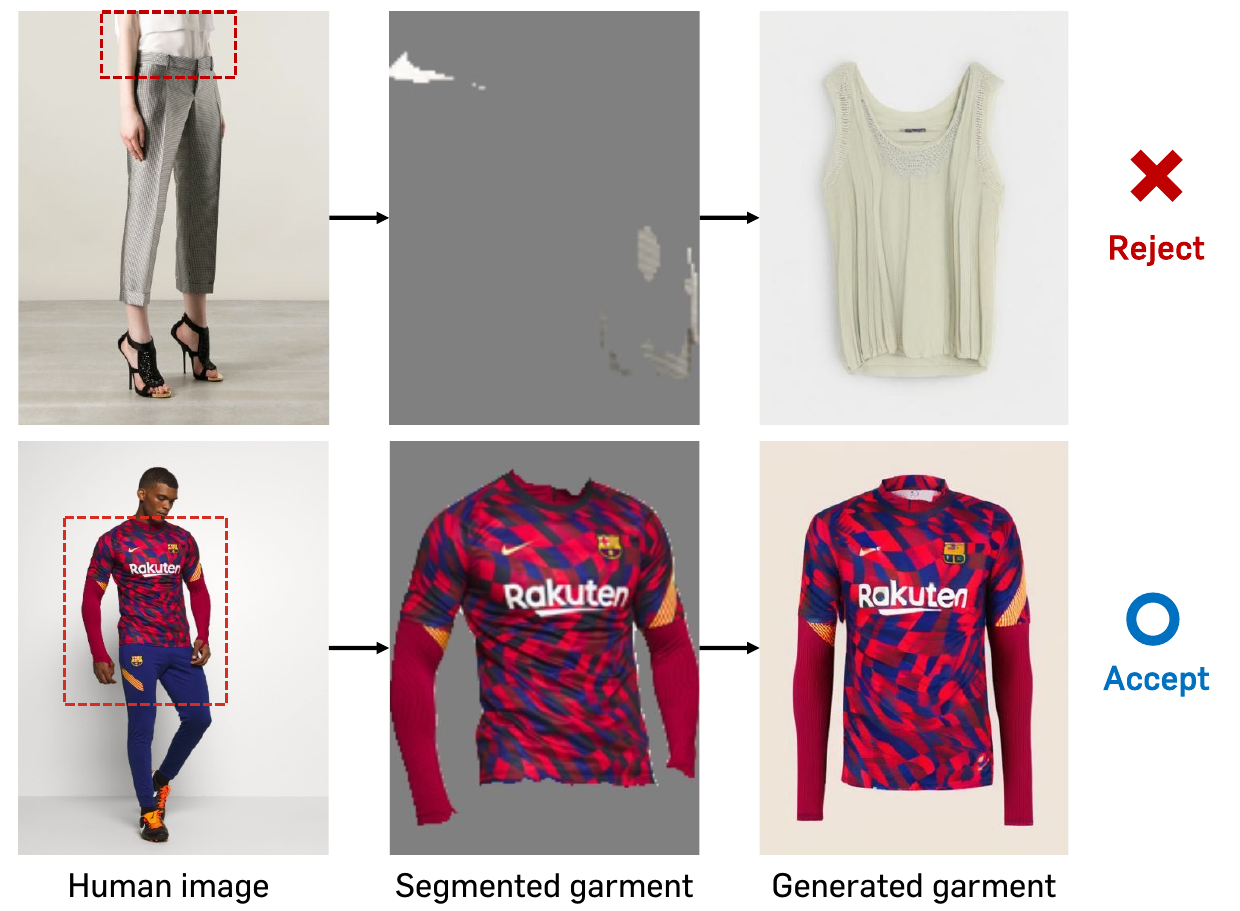}    
\captionof{figure}{\textbf{Examples of high\&low-quality generated garments.} When human parsing results are not precise, the decomposition network struggles to generate product garment images accurately, resulting in low-quality garment images. We filter out these cases.}
\vspace{-5pt}
\label{fig:lowq}
\end{figure}

\vspace{0.02in}
\noindent\textbf{Synthetic data generation with filtering.}\label{sec:filter}
After training the decomposition module, one can use it for extracting all of the reference images $\mathbf{X} = \{\rvx_1, \ldots, \rvx_N \}$ from each human image $\rvy$. It results in a synthetic dataset $\tilde{\mathcal{D}}$, which can be used for the conditional generative model $g_\phi$ for our goal of controllable generation. However, we find that the decomposition network $f_{\phi}$ sometimes generates low-quality reference images, especially when the prediction results from the parsing model $S$ are incorrect, which might harm the performance of $g_\phi$ (see Fig.~\ref{fig:lowq}).

Thus, we introduce a simple filtering strategy to improve the quality of our synthetic dataset $\tilde{\mathcal{D}}$. Specifically, we measure the image similarity score between the generated garment image $\tilde{\rvx}=f_{\phi}(\rvy,\rvm)$ and the segmentation results $\rvx^s$. We discard pair sets if any garment in the set has a similarity score below the threshold value $\tau > 0$, namely:
\begin{equation}
    d(\rvx^s,\tilde{\rvx} ) < \tau
  \label{eq:filter}
\end{equation}
For the scoring function for image similarity, we empirically find that dreamsim~\cite{fu2024dreamsim} aligns the most with human perception (See Appendix~\ref{sec:appendix_data} for details).

\subsection{Composition module} \label{sec:comp}
Our composition module consists of two diffusion models: one for a generation and the other one for an image encoder, denoted by $g_{\theta^{-}}$ and $g_{\theta}$, respectively. Both networks are initialized with the same pre-trained T2I diffusion models, where we freeze $g_{\theta^{-}}$ used as a generator and only train the encoder network $g_{\theta}$ using the synthetic dataset $\tilde{\mathcal{D}}$. In particular, the encoder $g_{\theta}$ is used to provide conditioning of garments $\tilde{\mathbf{X}}$ to the generator $g_{\theta^{-}}$.

To condition $\tilde{\mathbf{X}}$ to the generation model $g_{\theta^{-}}$, we concatenate the key and value vectors in each self-attention layer computed with each $\tilde{\rvx} \in \tilde{\mathbf{X}}$ and corresponding category $\rvm_{\tilde{\rvx}}$ using the encoder model $g_{\theta}$. By doing so, generator $g_{\theta}$ can be conditioned on $\tilde{\mathbf{X}}$ through its attentions. In particular, query, key, and value vectors of each of the attention layer in $g_{\theta}$ are computed with the following vectors
\begin{gather}
    \text{query}:=\mathbf{h}_{\rvy},\quad\text{key, value}:=[\mathbf{h}_\mathbf{y}, \mathbf{h}_{\tilde{\rvx}_1}, \ldots, \mathbf{h}_{\tilde{\rvx}_N}],
 \end{gather}
where $\mathbf{h}_{\rvy}$ and $[\mathbf{h}_{\tilde{\rvx}_1}, \ldots, \mathbf{h}_{\tilde{\rvx}_N}]$ are hidden states before the self-attention layer computed with the generation model $g_{\theta^{-}}$ and the encoder model $g_{\theta}$, respectively. To compute each $\mathbf{h}_{\tilde{\rvx}}$ we provide the text caption ``A photo of \textit{\{category\}}'' to the encoder model $g_{\theta}$, where \textit{\{category\}} is a type of garment $\tilde{\rvx}$.

Thus, we fine-tune the encoder $g_{\theta}$ through the diffusion model objective of the generator $g_{\theta^{-}}$:
\begin{align}
    \mathcal{L}(\theta):= \mathbb{E} \Big[ \omega(t) \big|\big| g_{\theta^{-}}(\mathbf{y}_t;\mathbf{c},t, \tilde{\mathbf{X}}) -
    \boldsymbol{\epsilon} \big|\big|_2^2 \Big],
\end{align}
where we employ synthetic text description for human image generated by vision-language model~\cite{liu2023llava} for $\rvc$.

\section{Experiments}

\begin{figure*}[ht]
\includegraphics[width=\textwidth]{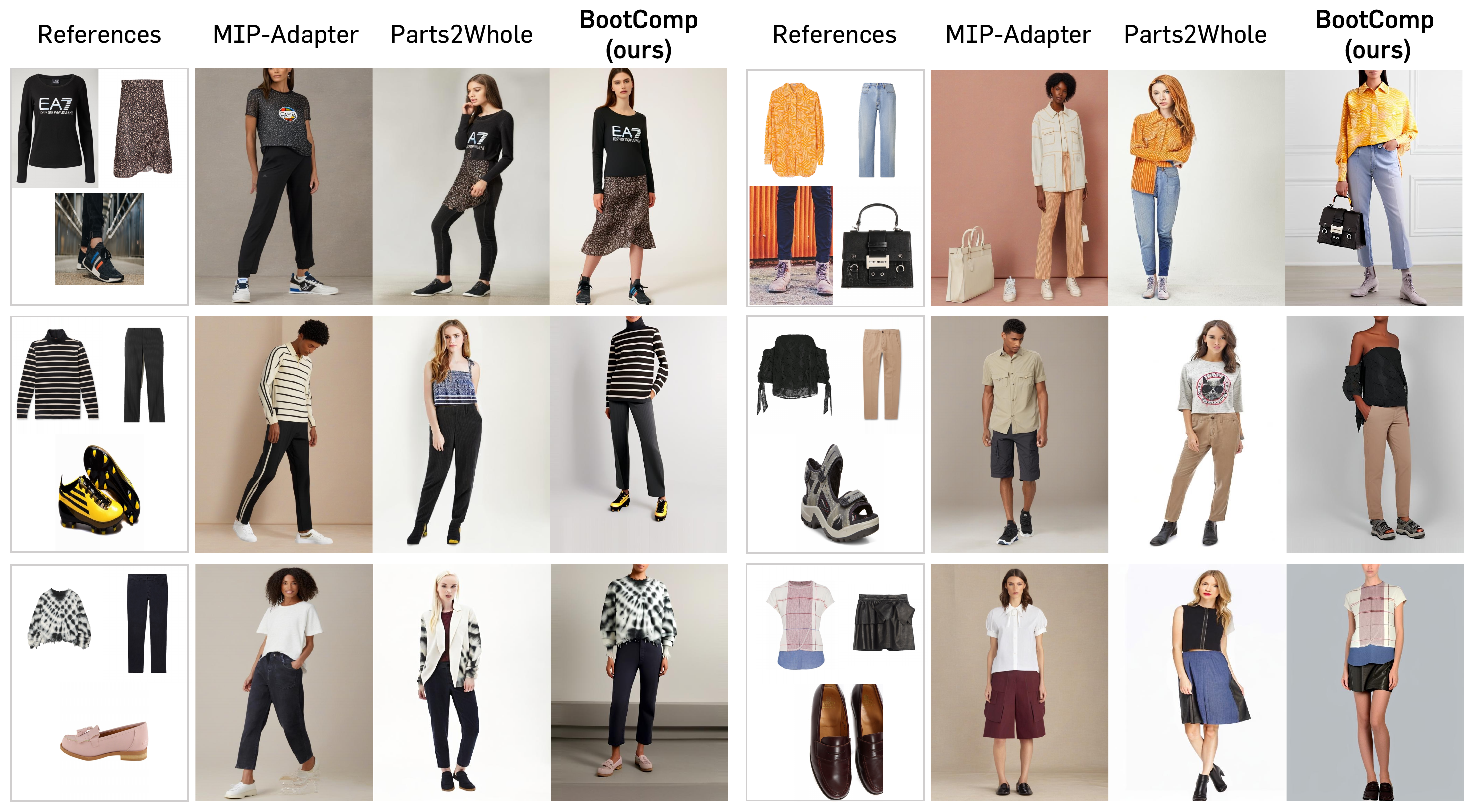}    
\captionof{figure}{\textbf{Qualitative comparison of human image generation with multiple garments.} \sname~generates realistic human images with multiple reference garments even with non-straightforward combinations of garments without losing details of each reference. For example, Parts2Whole replaces reference soccer cleats with stilettos, while ours accurately generates each reference (left, middle row).}
\label{fig:vis_comparison}
\end{figure*}

We validate the effectiveness of \sname~and the effect of the proposed components through extensive experiments. In particular, we investigate the following questions:
\begin{itemize}[leftmargin=5mm,itemsep=0mm]
    \item Can \sname~generate authentic human images wearing multiple garments while preserving details? (\cref{tab:main1}, \cref{fig:vis_comparison})
    \item Is our data generation pipeline effective and scalable, ensuring the model's performance? (\cref{tab:table_scale,tab:table_seg},~\cref{fig:compare_seg_qual})
    \item Can \sname~be  used for a wide range of downstream tasks? (Fig.~\ref{fig:applications})
\end{itemize}

\subsection{Experiment Setup}

We explain some important experimental setups in this section. We include more details in Appendix~\ref{sec:appendix_impl}.

\vspace{0.02in}
\noindent\textbf{Implementation details.}
We use Stable Diffusion XL (SDXL)~\cite{podell2023sdxl} for model initializations. We collect human-\textit{single} reference garment paired datasets from VITON-HD~\cite{choi2021viton}, DressCode~\cite{morelli2022dress} and LAION-Fashion~\cite{lepage2023lrvsf} for training the decomposition module. The dataset consists of 25,210 upper garments, 7,151 lower garments, 27,677 dresses, 5,675 bags, 1,599 shoes, 825 scarf, and 159 hats, resulting 68,296 single reference pairs on different categories. 
We train the decomposition module for 140K iterations with a total batch size of 32 on 4 H100 GPUs.
For the data generation phase, we process 240K human images obtained from VITON-HD, DressCode, LAION-Fashion, and DeepFashion~\cite{liuLQWTcvpr16DeepFashion} datasets, thereby collecting 240K paired data of human image and \textit{multiple} garment images at resolution 512$\times$384. 
We obtain and use 54K high-quality paired data after applying our filtering strategy with the threshold value $\tau=0.4$.
For the composition module, we train for 115K iterations with a total batch size of 48 on 8 H100 GPUs. For inference, we use the DDPM sampler \citep{ho2020denoising} with a sampling step of 50, where we apply classifier-free guidance (CFG; \citep{ho2022classifier}) with a guidance scale of $w=2.0$.

\vspace{0.02in}\noindent\textbf{Baselines.}
First, we consider MIP-Adapter~\cite{huang2024resolving} as baselines, which is a recent generic controllable generation method with multiple conditions. We also compare \sname~with FromParts2Whole~\cite{huang2024parts2whole}, the most relevant baseline for our task that aims for controllable human image generation with multiple reference garments.
We use the official model parameters from their official implementations. We employ ``A model wearing upper garment and lower garment and shoes'' as the text prompt to both models.

\vspace{0.02in}
\noindent\textbf{Evaluation metric.}
We report Frenchét Inception Distance (FID)~\citep{heusel2017gans}, MP-LPIPS \citep{chen2024magic}, and two different image similarities metrics~\citep{wang2024ms,huang2024resolving} using DINOv2~\citep{oquab2024dinov} (DINO and M-DINO). First, we use the FID score to measure the fidelity of generated human images, \ie, whether multiple garments are harmonized in the generated images. Next, we employ MP-LPIPS to evaluate the consistency of the target image to the source ground-truth garment. Finally, DINO and M-DINO measure the semantic similarity between each reference garment image and the respective garment present in the generated human image. 

\vspace{0.02in}\noindent\textbf{Evaluation datasets.}
We manually collect a dataset for evaluation as there are no common datasets for evaluating controllable human image generation. To evaluate MP-LPIPS, DINO, and M-DINO, we curate 5,000 garment image sets of three representative garment categories for human images (upper and lower garments and shoes). We randomly take upper and lower garment images from the test dataset of DressCode~\cite{morelli2022dress} dataset and shoe images from a public dataset.\footnote{\href{https://www.kaggle.com/datasets/noobyogi0100/shoe-dataset}{https://www.kaggle.com/datasets/noobyogi0100/shoe-dataset}}
Next, for evaluation using FID, we gather 30,000 human images wearing various garments in different poses from the test dataset of DressCode, VITON-HD, and Deepfashion to use them as reference image sets.

\begin{table}[t]
    \centering \small
    \caption{\textbf{Quantitative comparisons.} We compare \sname~with baselines on \sethlcolor{Peach!15}\hl{garment similarity} and \sethlcolor{YellowGreen!15}\hl{image fidelity}. We see that \sname~outperforms other methods, preserving fine-details of garments and naturally generating human images.
    }
    \setlength\tabcolsep{1pt}
    \begin{tabular}{@{}l cccc @{}}
    \toprule
    Method & \cellcolor{Peach!15} MP-LPIPS\,$\downarrow$ &\cellcolor{Peach!15} DINO\,$\uparrow$ &  \cellcolor{Peach!15} M-DINO\,$\uparrow$ &  \cellcolor{YellowGreen!15}FID\,$\downarrow$ \\
    \midrule
    MIP-Adapter~\citep{huang2024resolving} &0.276  &0.308 & 0.025 & 59.99 \\
    Parts2Whole~\citep{huang2024parts2whole} &0.267  &  0.362 &0.036  & 28.39  \\
    \textbf{\sname~(ours)} & \textbf{0.187} & \textbf{ 0.379} &\textbf{0.046}  & \textbf{27.63} \\
    \bottomrule
    \end{tabular}
    \label{tab:main1}
\end{table}

\subsection{Results}
\textbf{Qualitative results.}
We provide qualitative comparisons of our method (\sname) with other baseline methods in \cref{fig:vis_comparison}. As shown in this figure, \sname~generates more realistic human images in various poses, faithfully preserving details of reference garment images, while other methods often generate human images wearing garments inconsistent with the references. Moreover, this result shows that \sname~generates creative combinations of garments. For instance, in the first example of the second row, \sname~generates a human image with uncommon combination of garments (\eg, trousers with soccer cleats) but Parts2Whole or MIP-Adapter fails to achieve this: they either undesirably replace the cleats to trousers or struggle with generating high-fidelity garments (respectively). We provide more visualizations in Appendix~\ref{sec:appendix_res}.

\begin{figure*}[t]
\includegraphics[width=\textwidth]{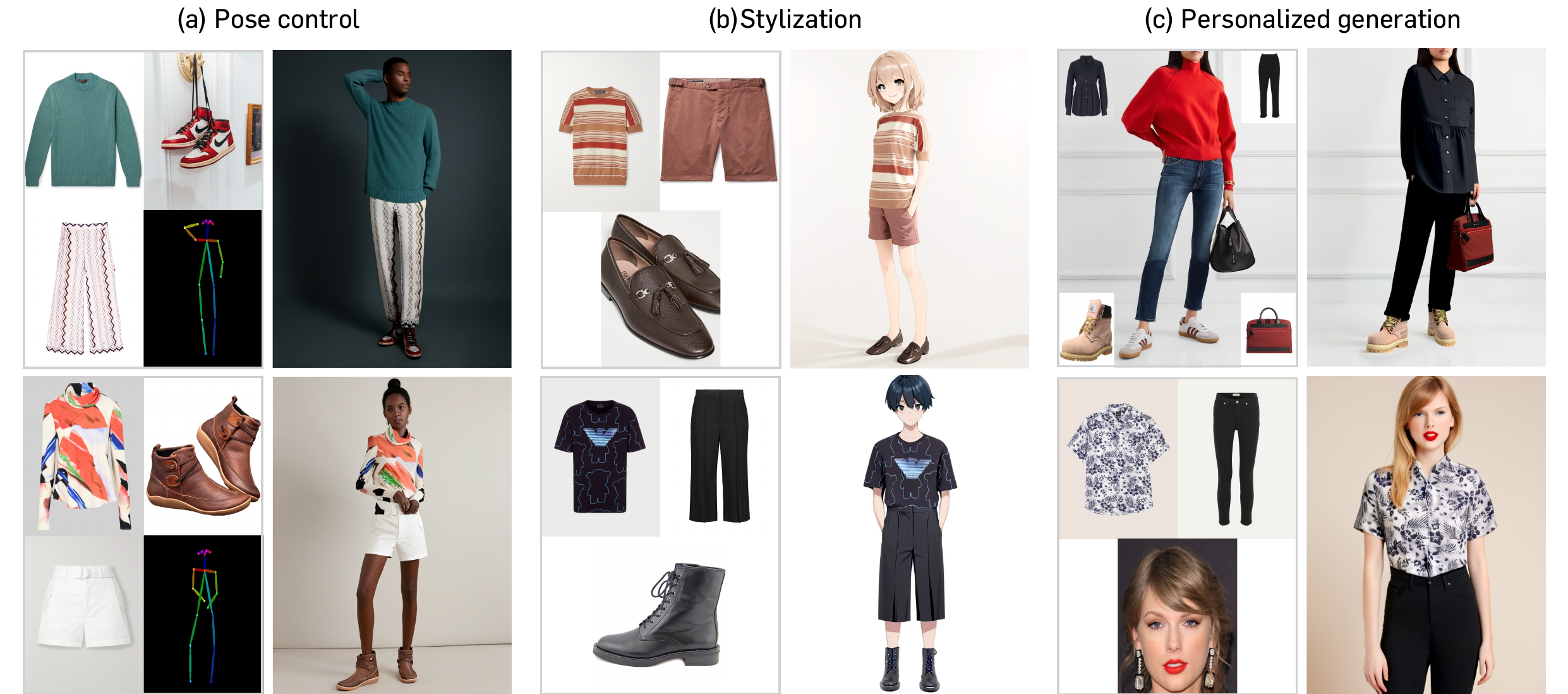}    
\captionof{figure}{\textbf{More applications of \sname.} We showcase the extensive applications of our method, \sname. \sname~creates human images by controlling the (a) poses and (b) styles of the generated human images. \sname~also enables (c) personalized human image generation by taking user's images as conditions (\eg, face, full body).}
\label{fig:applications}
\vspace{-0.1in}
\end{figure*}

\vspace{0.02in}
\noindent\textbf{Quantitative results.}
 We report quantitative evaluation results of \sname~and baselines in \cref{tab:main1}. \sname~outperforms both MIP-Adapter and Parts2Whole across all of four evaluate metrics. In particular, \sname~achieves a 30\% improvement in MP-LPIPS score over the baselines, demonstrating its effectiveness in preserving garment details. Moreover, \sname~shows its capabilities in authentic image generation for human images, as indicated by a better FID values than baselines.

\vspace{0.02in}
\noindent\textbf{More applications.}
In \cref{fig:applications}, we apply \sname~to several downstream tasks and visualize their results.
First, we show that \sname~can generate human images conditioned on the pose. In \cref{fig:applications} (a), \sname~generates human images in diverse poses following the extra conditions even with reference garments of intricate patterns, demonstrating its generalization capability.
We also show that \sname~can generate human images with different stylizations such as cartoons in \cref{fig:applications} (b). 
Finally, we show that \sname~can be used for personalized human image generation such as virtual try-on, \ie, changing garments on a given human image to reference garments. In \cref{fig:applications} (c), \sname~replaces garments on a given human image with the reference garment images and enables personalized generation conditioning face image.

Note that this can be done without any additional task-specific fine-tuning as we freeze the generator in the composition module during training. This enables \sname to be easily integrated with other modules, \eg, IP-Adapter~\cite{ye2023ip-adapter} or ControlNet~\cite{zhang2023adding}, that provides controllability with additional condition inputs. We provide additional generation results for each application in Appendix~\ref{sec:appendix_res}.

\begin{figure}[t]
\includegraphics[width=\linewidth]{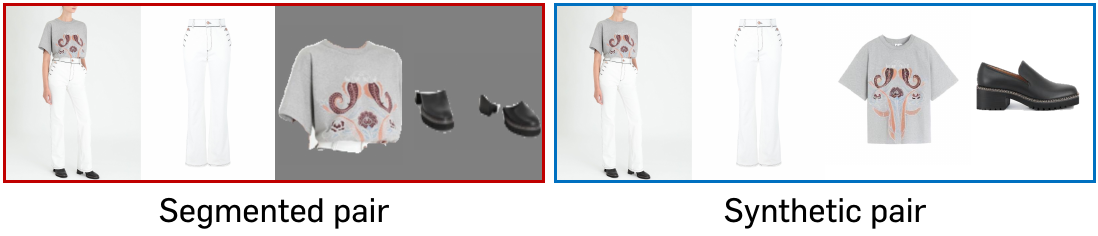}    
\captionof{figure}{\textbf{Visualization of segmented paired data and our synthetic paired data.}
We provide a visual comparison between segmented and synthetic pairs. Given a single garment and a human image pair, we segment out other garments from the human image in the segmented paired data.
}
\label{fig:compare_seg}
\end{figure}

\begin{table}[t]
    \centering\small
    \caption{\textbf{Comparison on dataset construction methods.} The model trained on the segmented paired dataset shows worse performance compared to one trained on our synthetic paired dataset both in \sethlcolor{Peach!15} \hl{garment similarity} and \sethlcolor{YellowGreen!15} \hl{image fidelity}.}
    \vspace{-0.05in}
    \setlength\tabcolsep{2pt}
    \begin{tabular}{@{}l cccc @{}}
    \toprule
    Dataset &  \cellcolor{Peach!15} MP-LPIPS\,$\downarrow$ & \cellcolor{Peach!15} DINO\,$\uparrow $ & \cellcolor{Peach!15} M-DINO\,$\uparrow$ & \cellcolor{YellowGreen!15} FID\,$\downarrow$ \\
    \midrule
    Segmented &0.374 &0.284  &0.025 & 59.27 \\
    \textbf{Synthetic} &\textbf{0.197}   &\textbf{0.365}   &\textbf{0.043}  & \textbf{29.41} \\
    \bottomrule
    \end{tabular}
    \label{tab:table_seg}

\vspace{-10pt}
\end{table}

\subsection{Analysis and ablation studies}
Finally, we conduct several analyses on synthetic data to validate our data generation pipeline, including its scalability and the impact compared with a na\"ive use of a segmented paired dataset. To reduce the computation cost, we use Stable Diffusion v1.5 for all analyses while we strictly follow the other setups used in the main experiments.

\begin{figure}[t]
\includegraphics[width=\linewidth]{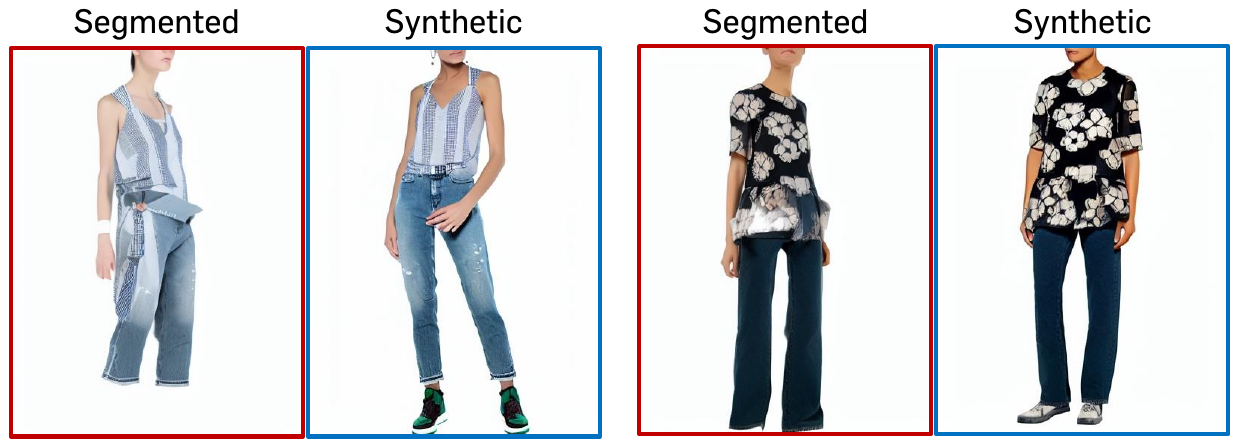}    
\captionof{figure}{\textbf{Visual comparison on data construction methods.}
Visual comparison between generated human images where each model is trained on segmented and synthetic pairs. The model trained on segmented pair data struggles to generate naturally harmonized human images (red).
}
\label{fig:compare_seg_qual}
\end{figure}

\noindent\textbf{Effect of data generation.}
We first show the effect of our data generation scheme. We demonstrate this by constructing a dataset by segmenting out all garment images from the human except the given one in the dataset (see  \cref{fig:compare_seg}), and train the composition module on this dataset. As shown in \cref{tab:table_seg}, the model trained on the segmented paired dataset achieves worse performance across all evaluation metrics. Also, \cref{fig:compare_seg_qual} visualizes undesirable generated images by the model trained on the segmented dataset.
This indicates the model struggles to generate desirable human images, highlighting the effectiveness of our data generation scheme.

\vspace{0.02in}
\noindent\textbf{Scalability of the data generation scheme.}
Next, we investigate the scalability of our data generation scheme by exploring the effect of dataset size to the performance. We observe that using a larger dataset for training always improves the model's performance in both garment fidelity and image fidelity, as shown in \cref{tab:table_scale}. 

\begin{table}[t]
    \centering\small
    \caption{\textbf{Comparison on dataset scale.} Training with a larger datatset (after filtered) improves the model's overall performance in both \sethlcolor{Peach!15} \hl{garment similarity} and \sethlcolor{YellowGreen!15} \hl{image fidelity}. 
    }
    \vspace{-0.05in}
    \setlength\tabcolsep{2pt}
    \begin{tabular}{@{}l ccc @{}}
    \toprule
    Dataset size & \cellcolor{Peach!15} DINO\,$\uparrow $ & \cellcolor{Peach!15} M-DINO\,$\uparrow$ &  \cellcolor{YellowGreen!15} FID\,$\downarrow$ \\
    \midrule
    5K  & 0.337 & 0.248  & 34.15 \\
    15K & 0.338 & 0.251  & 32.32  \\
    30K & 0.344 & 0.261  &  26.99 \\
    \textbf{50K} & \textbf{0.360} & \textbf{0.285}  & \textbf{25.88}  \\
    \bottomrule
    \end{tabular}
    \label{tab:table_scale}

\end{table}

\begin{table}[h!]
    \centering\small
    \caption{\textbf{Ablation study for threshold value $\tau$ on filtering.} The data quality improves with a stricter threshold value, leading to better performance. We adopt $\tau=0.4$ when applying the filtering.}
    \vspace{-0.05in}
    \begin{tabular}{cccccc}
        \toprule
        $\tau$ & 0.4 & 0.5 & 0.6 & 0.7 & 1.0 \\
        \midrule
        DINO$\uparrow$ & \textbf{0.360} & 0.347 & 0.343 & 0.342 & 0.338 \\
        \bottomrule
    \end{tabular}
    \label{table:ablation}
\end{table}
\vspace{0.02in}\noindent\textbf{Ablation study: threshold value $\tau$.}
Finally, we conduct an ablation study on the threshold value $\tau$ used in our dataset filtering strategy. In Table.~\ref{table:ablation}, we report similarity score (DINO) of the models trained with  different datasets by varying values of $\tau$ from 0.4 to 1.0, where 1.0 indicates no filtering is applied. We observe that more strict data filtering can provide more performance gain to the model.

\section{Related Work}
\label{sec:Related}

\textbf{Controllable image generation.}
In addition to using text prompts as conditions, recent works have attempted to improve the controllability of text-to-image (T2I) diffusion models by incorporating additional inputs (\eg, images). 
In particular, many works focus on generating images that preserve the identity of subjects in the source image by proposing additional modules to the model~\citep{ye2023ip-adapter, kosmos-g, li2023blip}. Despite their effort, they have struggled to generalize with multiple subjects and suffer from several issues, such as subject blending. To mitigate this issue, several approaches such as MS-Diffusion~\citep{wang2024ms} and FastComposer~\citep{xiao2023fastcomposer} introduce an additional regional information for each subject. Our framework also tries to improve image generation with multiple subjects, but we focus on human image generation and propose a novel data generation pipeline to improve the quality.

\vspace{0.02in}\noindent\textbf{Virtual try-on.}
Inspired by the great progress of T2I diffusion models, recent works have explored their application to various tasks on fashion domain such as virtual try-on~\cite{Zhu_2023_CVPR_tryondiffusion,kim2024stableviton,Zhu_2024_CVPR_mmvto,PARK2025104259,choi2024improving,chen2024wear,zhang2024mmtryon} and virtual dressing~\cite{shen2024imagdressing,chen2024magic}. However, most of them are limited to single-garment based generation as they rely on existing public datasets~\cite{choi2021viton,morelli2022dress} consisting of single-paired data. While several works~\cite{Zhu_2024_CVPR_mmvto,PARK2025104259,zhang2024mmtryon} address multi-garment virtual try-on, they depend on proprietary datasets, which limits scalability and its capability to support a few garment categories. Our data generation pipeline tackles this data acquisition bottleneck and supports multi-garment based generation with a wide range of categories.

\vspace{0.02in}\noindent\textbf{Improving diffusion models with self-data generation.}
Recent works have tried to improve the performance of the pre-trained model itself on the specific tasks~\cite{brooks2023instructpix2pix, jang2024identity, zeng2024jedi, gal2024lcm, winter2024objectdrop} using generated image data from the same model. For example, JeDi \citep{zeng2024jedi} generates same-subject images using LLMs and pretrained T2I diffusion models. They are used to fine-tune T2I diffusion models for personalized generation \citep{ruiz2022dreambooth} without additional tuning at inference.

However, these approaches are not suitable for the case of images with multiple subjects, such as controllable human generation, as most T2I diffusion models still lack the capability to accurately generate images with multiple subjects \citep{nie2024compositional,li2023gligen}. As a result, synthetic image data with multiple subjects generated with T2I models often exhibit low-quality results, and thus fine-tuning with this dataset does not lead to the improvement. Thus, rather than generating multi-subject images from T2I models, existing approaches have curated data through a segmentation from the multi-subject images~\citep{huang2024parts2whole}. 
However, these models suffer from the copy-and-paste and subject inconsistency problems. Our method bridges the former and latter approaches to improve data quality used for controllable human generation.

\section{Conclusion}
\label{sec:Conclusion}

In this paper, we present \sname, a novel framework for controllable human image generation with multiple garments given as image conditions. Our pipelines for synthetic paired data generation and controllable generation enabled creating human images wearing multiple reference garments. We show the broad applicability of \sname~by adapting it to various types of tasks in the fashion domain.

\section*{Acknowledgement}
This work was supported by Institute for Information \& communications Technology Promotion (IITP) grant funded by the Korea government (MSIT) (No.RS-2019-II190075 Artificial Intelligence Graduate School Program(KAIST); No.RS-2021-II212068, Artificial Intelligence Innovation Hub).

\clearpage
{
    \small
    \bibliographystyle{ieeenat_fullname}
    \bibliography{main}

}


\appendix
\clearpage
\setcounter{page}{1}
\maketitlesupplementary

\section{Implementation Details} \label{sec:appendix_impl}
\subsection{Training and Inference}
We train our decomposition module on 68,296 pairs of a human image and a single reference garment image at 512$\times$384 resolutions with a fixed learning rate of 1e-5 using Adam optimizer~\cite{kingma2014adam}. We train for 140K iterations with a total batch size of 32 using 4 H100 GPUs.

For the composition module, we train on 54K pairs of a human image and multiple reference garment images at 768$\times$576 resolution with a fixed learning rate of 1e-5 and Adam optimizer. We train for 115K iterations with a total batch size of 48 using 8 H100 GPUs.

During the inference, we generate images using the DDPM~\citep{ho2020denoising} sampler with 50 denoising steps. We apply classifier-free guidance (CFG) \citep{ho2022classifier} with the text conditioning $\rvc$ and garment image conditioning $\mathbf{g}$ as follows:
\begin{align*}    
    \hat{\boldsymbol{\epsilon}}_\theta(\mathbf{x}_t;\mathbf{g},\rvc,t) = w\cdot (\boldsymbol{\epsilon}_\theta(\mathbf{x}_t;\mathbf{g},\rvc,t) - \boldsymbol{\epsilon}_\theta (\mathbf{x}_t;t)) + \boldsymbol{\epsilon}_\theta(\mathbf{x}_t;t)\text{,}
\label{guidance}
\end{align*}
where $\boldsymbol{\epsilon}_\theta(\mathbf{x}_t;\rvc,\mathbf{g},t)$ denotes noise prediction output with text and garment image conditions, and $\boldsymbol{\epsilon}_\theta(\mathbf{x}_t;t)$ denotes the unconditional noise prediction output. We use a guidance scale of $w=2.0$ for sampling.

\begin{figure}[!t]
\includegraphics[width=\linewidth]{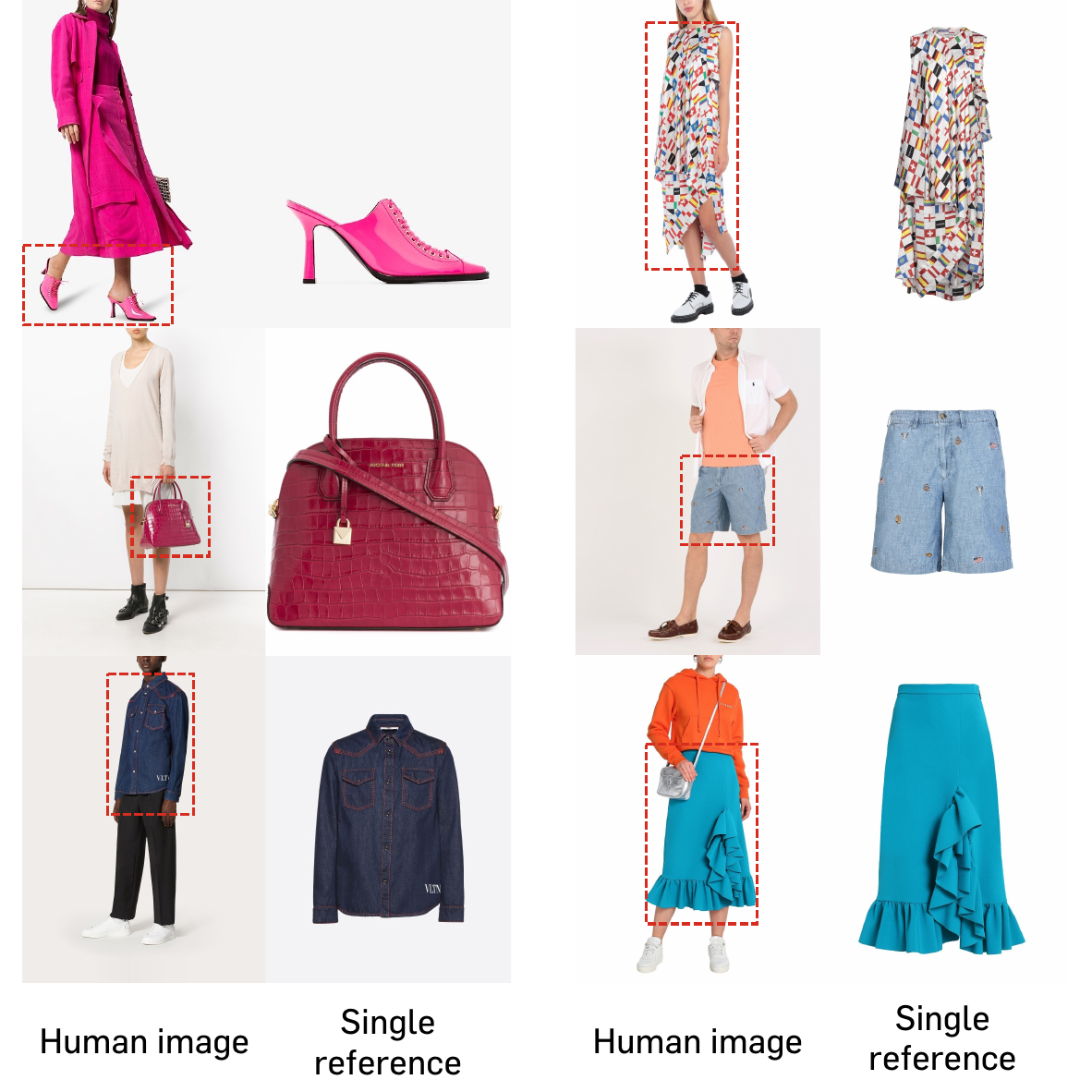}    
\captionof{figure}{\textbf{Examples of training data for decomposition module.} We collect pairs of a human image and a single reference garment image from public datasets including VITON-HD, DressCode, and LAION-Fashion. It consists of various garments in different categories, \eg, shirts, pants, shoes and bags \etc.
}
\vspace{-15pt}
\label{fig:singlepair_eg}
\end{figure}

\subsection{Single reference Paired Dataset}
To train the decomposition network, we collect pairs of a human image and a single reference garment image from VITON-HD, DressCode, and LAION-Fashion datasets. Specifically, we gather 11,647 upper garments and human images from the training dataset on VITON-HD. We also collect 13,563 upper garments, 7,151 lower garments, 27,677 dresses paired with human images from DressCode. For LAION-Fashion dataset, since it consists of single reference pairs without categorical information, we use CLIP~\cite{radford2021learning} model to classify the garment image. We define 19 different garment category texts and match the garment image with the category text of the highest similarity score, resulting in 5,675 bags and 1,599 shoes, 826 scarf, and 159 hats in the training data. We provide examples of collected single reference garment and human image pairs in~\cref{fig:singlepair_eg}.

\subsection{Dual-Condition Classifier-free Guidance}
Since we have dual conditions of text condition $\rvc$ and garment image condition $\mathbf{g}$, one can apply classifier-free guidance for two conditions following~\cite{brooks2023instructpix2pix}. Formally:
\begin{align*}    
\hat{\boldsymbol{\epsilon}}_\theta(\mathbf{x}_t;\mathbf{g},\rvc,t) = & \, w_c\cdot (\boldsymbol{\epsilon}_\theta(\mathbf{x}_t;\mathbf{g},\rvc,t) - \boldsymbol{\epsilon}_\theta (\mathbf{x}_t;\mathbf{g},t)) \\
& + 
    w_g\cdot (\boldsymbol{\epsilon}_\theta(\mathbf{x}_t;\mathbf{g},t) - \boldsymbol{\epsilon}_\theta (\mathbf{x}_t;t)) \\
  &  +    
    \boldsymbol{\epsilon}_\theta(\mathbf{x}_t;t)\text{,}
\end{align*}
where $w_c>0$ and $w_g>0$ denotes a guidance scale for text conditioning and  garment image conditioning, respectively. Increasing $w_g$ encourages generated images to more similar to the reference garment images, and increasing $w_c$ guides the generated images to better align with the given text prompt. While we adopt $w_g=2.0$ and $w_c=2.0$ for all experiments, users can adjust the guidance values to customize the generated images according to their preferences.

\section{Synthetic Dataset Construction} \label{sec:appendix_data}
In this section, we provide a detailed explanation of the data curation process with visualizations. 

\begin{figure}[!bht]
\includegraphics[width=\linewidth]{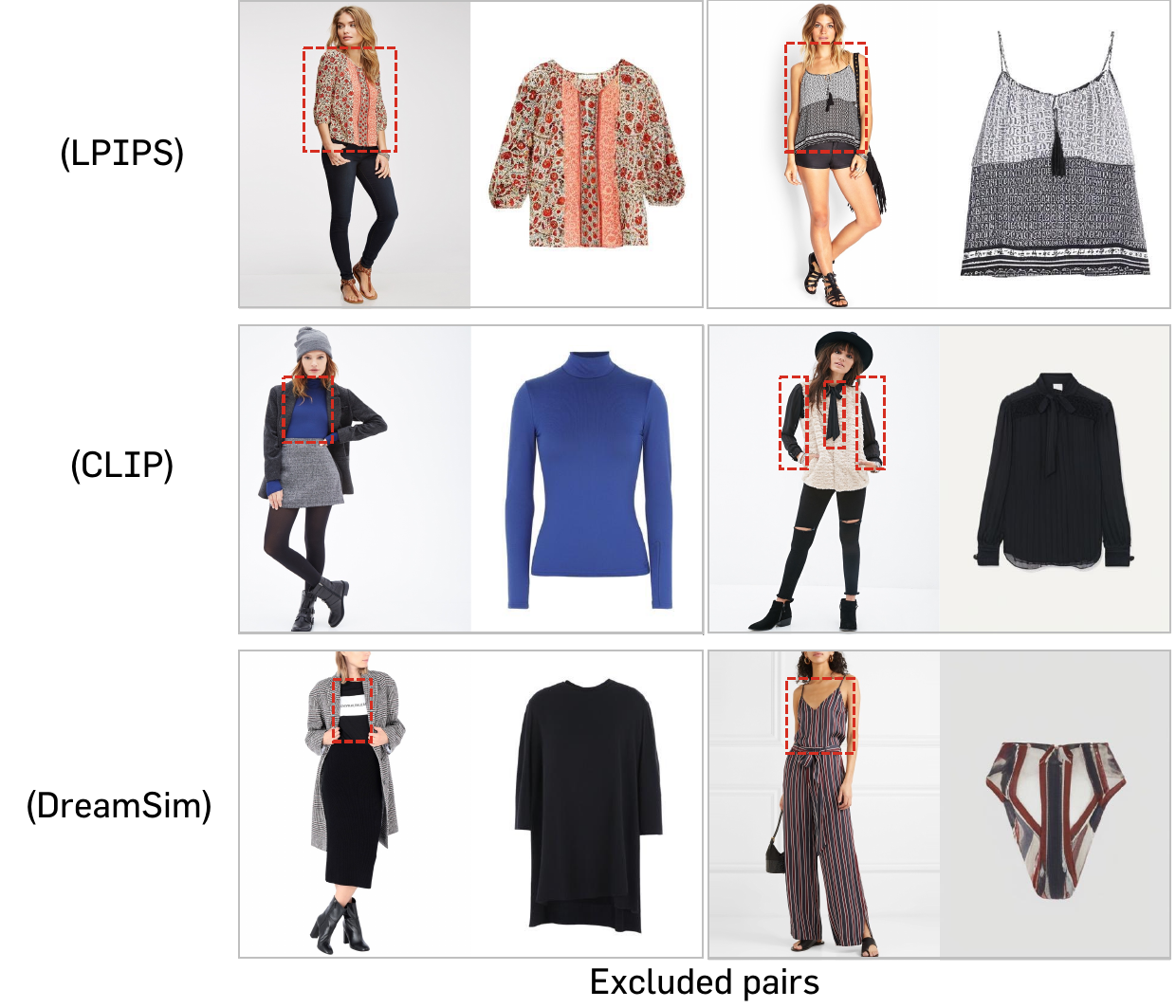}    
\captionof{figure}{\textbf{Examples of pairs filtered out by different similarity metrics.} We present examples of generated garment images and their corresponding human images that were excluded based on various image similarity metrics. Using LPIPS, garments with complicated patterns are filtered out, and using CLIP score, inner layer garments are filtered out even when they are considered identical in human perception. In contrast, DreamSim captures the distance between images in a way aligned with human perception, filtering out undesirable pairs.
}
\vspace{-5pt}
\label{fig:simscore_eg}
\end{figure}

\begin{figure}[ht]
\includegraphics[width=\linewidth]{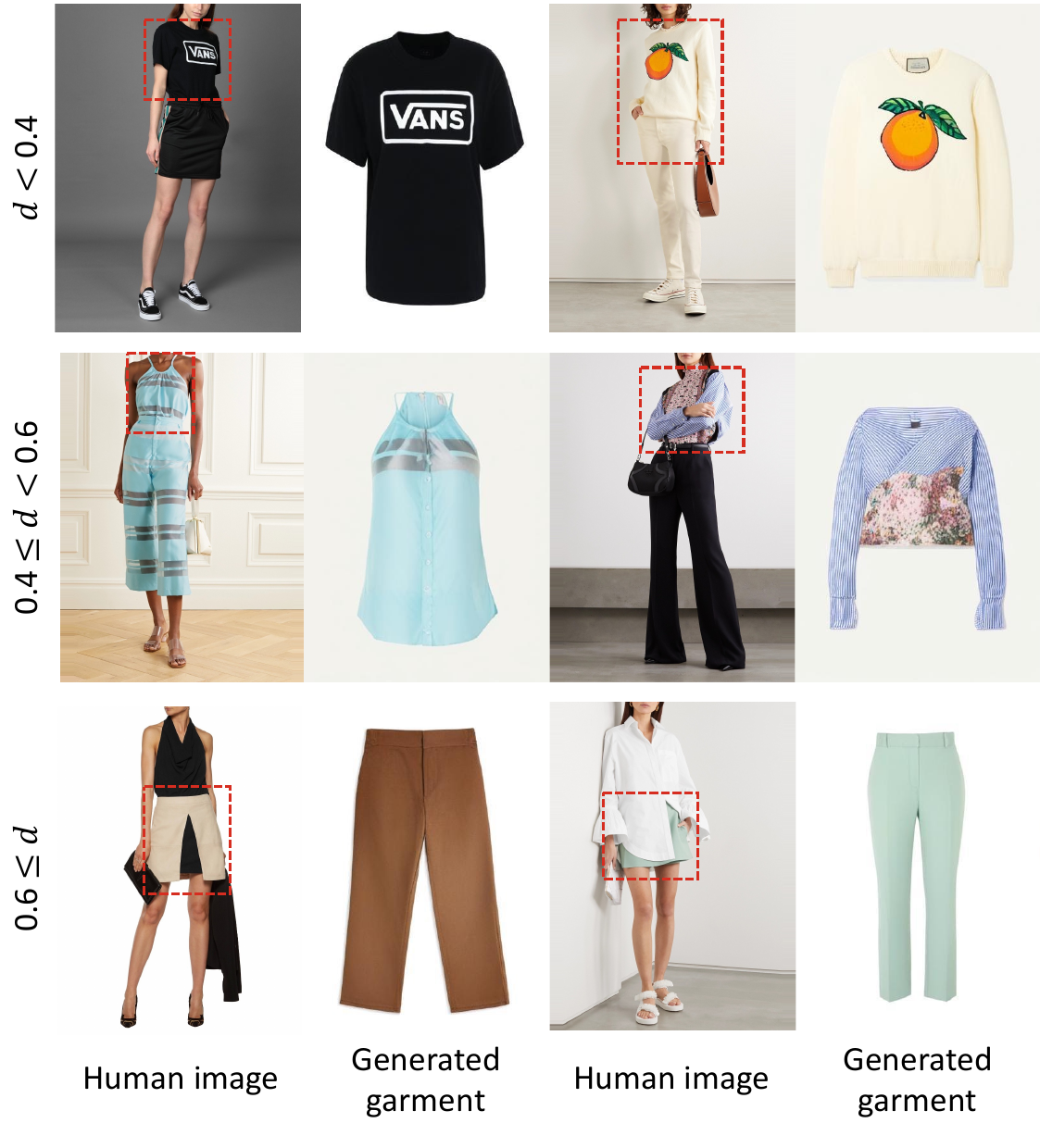}    
\captionof{figure}{\textbf{Examples of generated garment images with different image distance values.} We provide examples of generated garment images and corresponding human images, varying the distance values measured by DreamSim. With the distance value $d\geq0.4$, generated garments are inconsistent with the actual garment, while for $d<0.4$, the generated garments closely resemble the actual garment.}
\vspace{-10pt}
\label{fig:distance_eg}
\end{figure}

\subsection{Filtering Strategy}
As illustrated in~\ref{sec:filter}, we apply filtering on our synthetic paired data based on the image similarity between the segmented and generated garments. Among several possible metrics, we try LPIPS, CLIP score, and DreamSim, and empirically find that DreamSim aligns the most with human perception. As shown in~\cref{fig:simscore_eg}, DreamSim can measure the similarity aligned with human perception and filters out undesirable samples while CLIP and LPIPS struggle.
For example, LPIPS determines that similar garments do not resemble each other, even if garment pairs look identical to humans, especially when they contain intricate patterns or stripes. Also, CLIP fails to identify the same garments, mainly when garments are inner layers under jackets, whereas DresmSim captures similarity in a way aligned with human perception, filtering out the undesirable pairs.

We adopt DreamSim for measuring the distance between segmented garments and generated garments. We visualize human images and generated garment images based on the image distance value in~\cref{fig:distance_eg}. With the distance value $d\geq0.6$, we observe that the generated garment is inconsistent with the garment on the human image, and with $0.4\leq d <0.6$, fine details are not fully preserved. On the other hand, with $d<0.4$, generated garments closely resemble the actual garments.

\begin{figure}[!th]
\includegraphics[width=\linewidth]{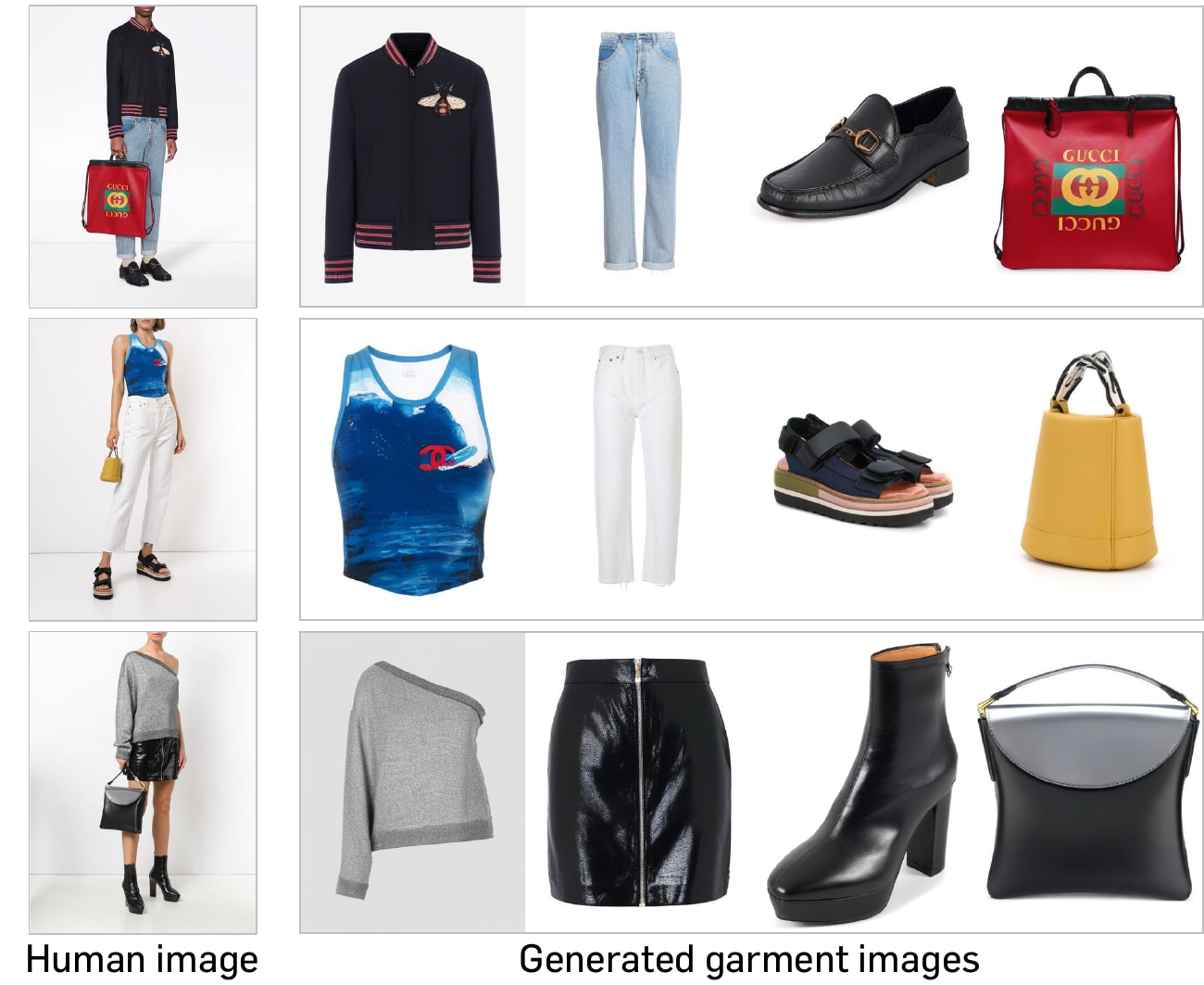}    
\captionof{figure}{\textbf{Examples of our synthetic paired data.} We visualize our synthetic pairs of a human image and multiple garment images. Our decomposition module generates high-quality garment images in product view on different categories including shirts, pants, shoes and bags.
}
\vspace{-5pt}
\label{fig:dataset_eg}
\end{figure}

\subsection{Synthetic Dataset Examples}
We provide visualizations of the synthetic paired dataset generated by our decomposition network in~\cref{fig:dataset_eg}. The synthetic dataset contains high-quality pairs of a human image and \textit{multiple} reference garments. The decomposition network can generate product garment images on different categories, even with challenging garments such as one-shoulder sweaters (Third-row in~\cref{fig:dataset_eg}).  

\begin{figure}[!th]
\includegraphics[width=\linewidth]{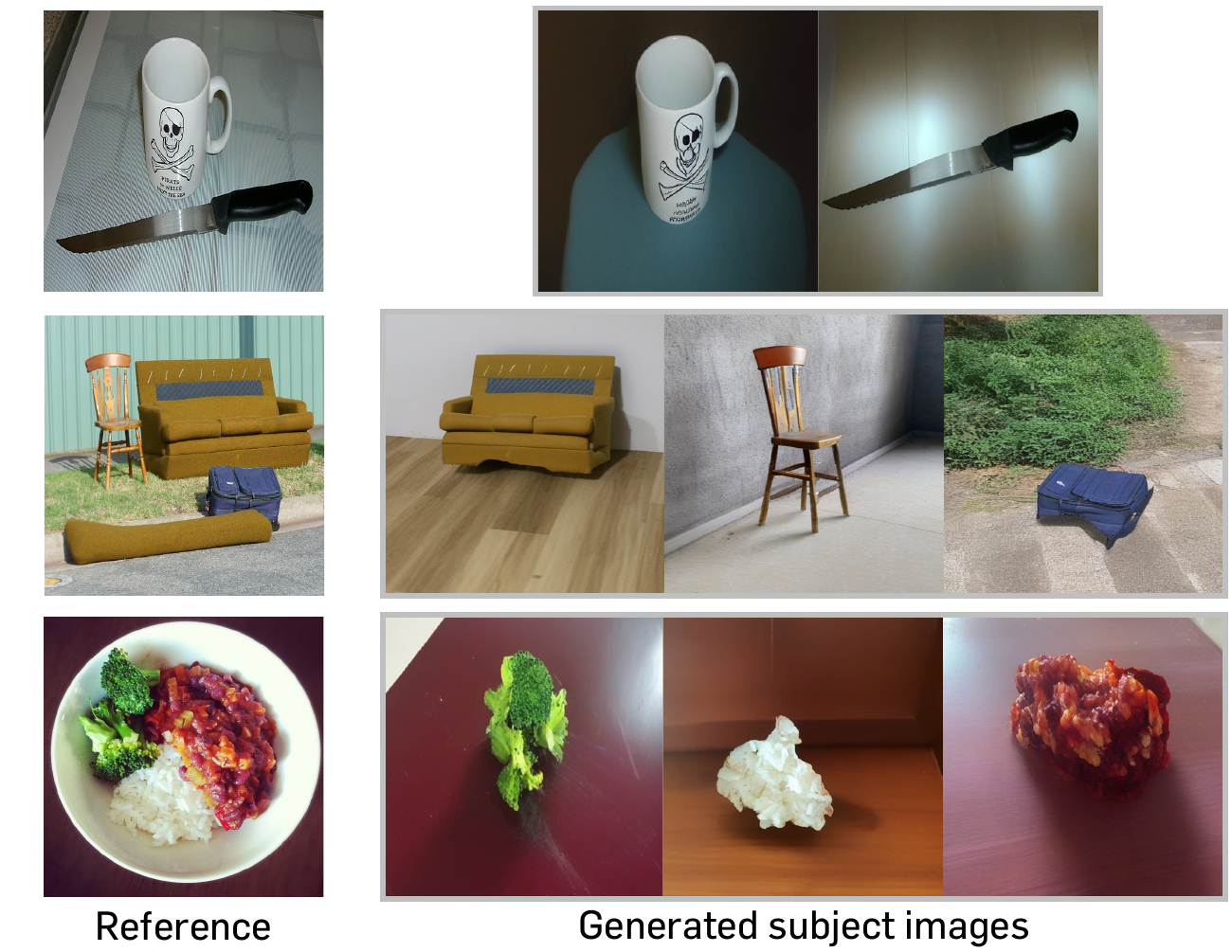}    
\captionof{figure}{\textbf{Examples of synthetic paired data generated by the decomposition module trained on MVImgNet~\cite{yu2023mvimgnet}.} We show the potential extension of our decomposition module to the general domain. Given an image containing common objects such as cups, chairs, and broccoli, the decomposition module generates each object in a different view, constructing paired data. Reference images are obtained from COCO~\cite{lin2014microsoft}.
}
\vspace{-15pt}
\label{fig:general}
\end{figure}
 
\section{Applications of Decomposition module} 
\label{sec:appendix_app}
In this section, we explore the potential applications of our decomposition module, including applying it on the general domain and using it as a multi-view image generator.

\subsection{Synthetic Paired Data on General Domain}
Recent work~\cite{xiao2024omnigen} demonstrates remarkable performance in diverse image generation tasks by leveraging large-scale paired data, underscoring the importance of paired datasets in image generation. We have demonstrated our decomposition module's capability to generate high-quality paired data in the fashion domain, and we further explore its potential for applicability to the general domain. Specifically, we train the decomposition network on MVImgNet~\cite{yu2023mvimgnet} dataset, which contains large-scale object images in multi-view from 238 classes. As shown in~\cref{fig:general}, the network decomposes each object in different views from reference images, demonstrating its potential for broader applications and inspiring future research.

\subsection{Multi-view Image Generator}
\begin{figure}[!h]
\includegraphics[width=\linewidth]{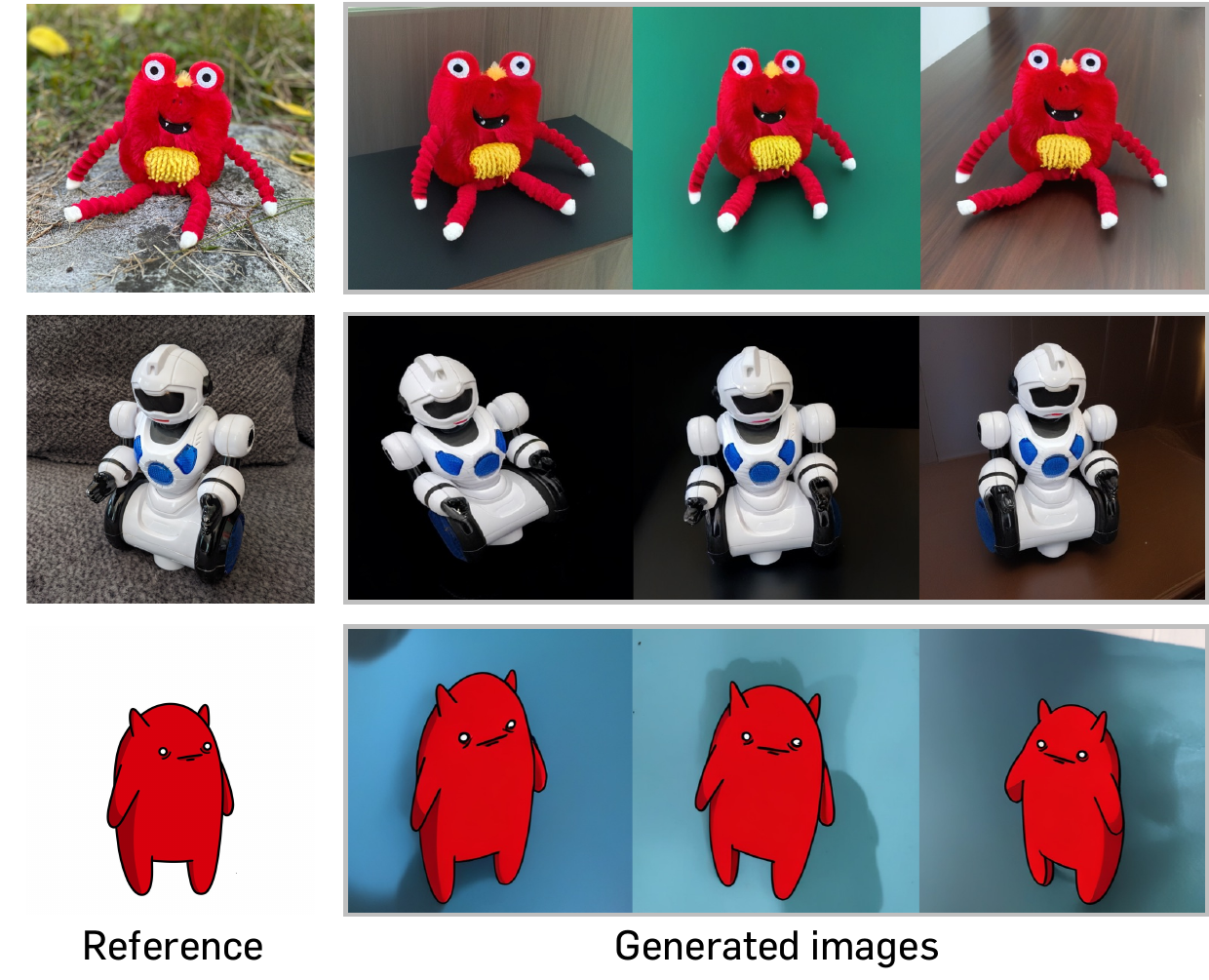}    
\captionof{figure}{\textbf{Examples of generated subjects in multi-view by the decomposition module trained on MVImgNet.} The decomposition module can serve as a multi-view generator for single-subject images. Subject images are from DreamBooth~\cite{ruiz2022dreambooth}. 
}
\vspace{-15pt}
\label{fig:multiview}
\end{figure}

We show that the decomposition network can be used as a multi-view image generator. By utilizing the decomposition network with segmented single-subject images, one can generate different views of the reference subject images while faithfully preserving their identity. In~\cref{fig:multiview}, we present multi-view images generated by the decomposition module using subject images obtained from DreamBooth~\cite{ruiz2022dreambooth}. These multi-view images can be utilized for various applications, such as data augmentation. 

\section{Additional Qualitative Results} \label{sec:appendix_res}
\begin{figure*}[t]
\includegraphics[width=\textwidth]{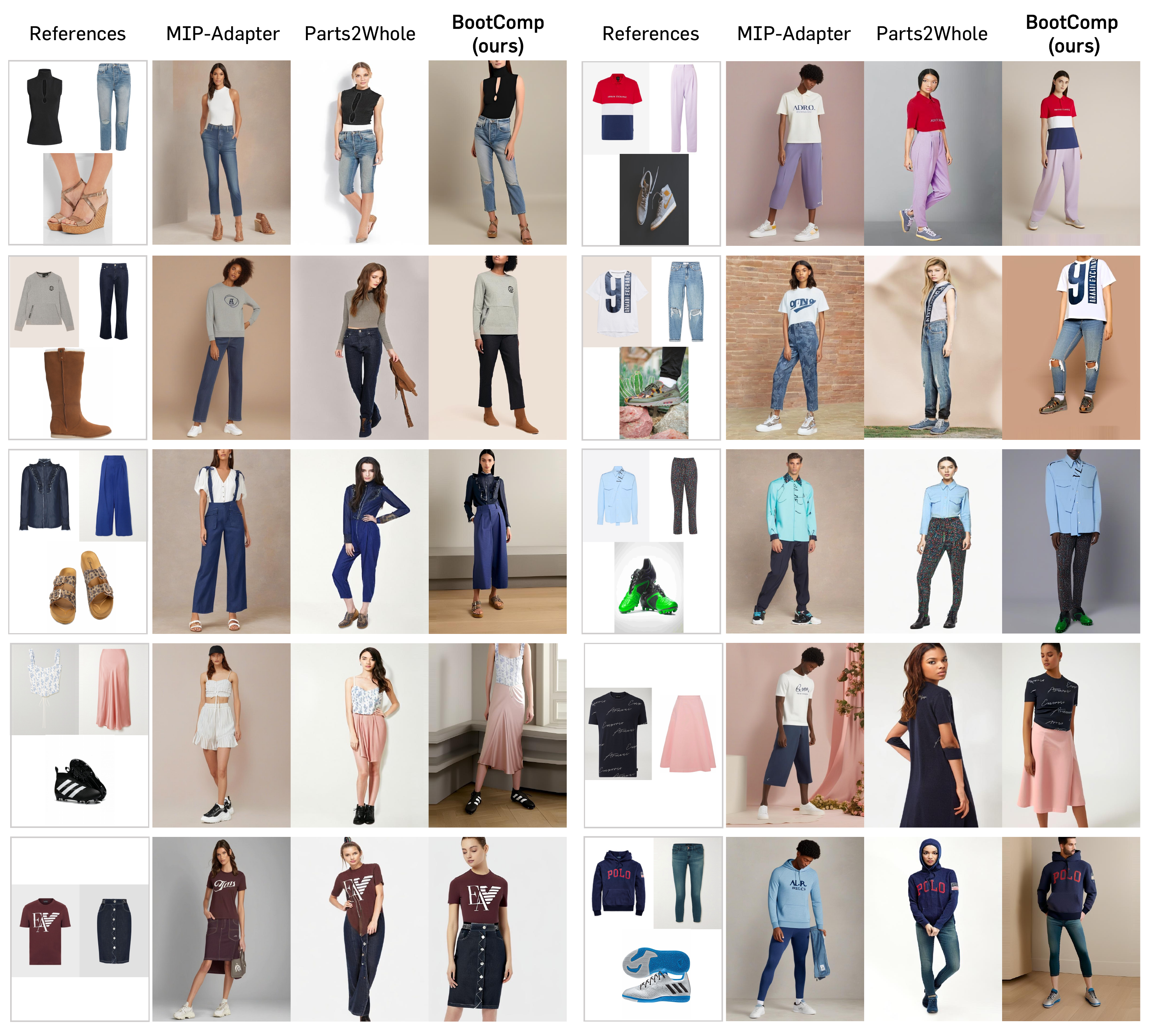}    
\captionof{figure}{\textbf{More qualitative comparisons.} \sname~generates realistic human images wearing multiple reference garments, faithfully preserving fine-details of each garment, while baselines often generate inconsistent garment images and blend reference garments.}
\label{fig:appendix_compare}
\vspace{-0.1in}
\end{figure*}

\begin{figure*}[t]
\includegraphics[width=\textwidth]{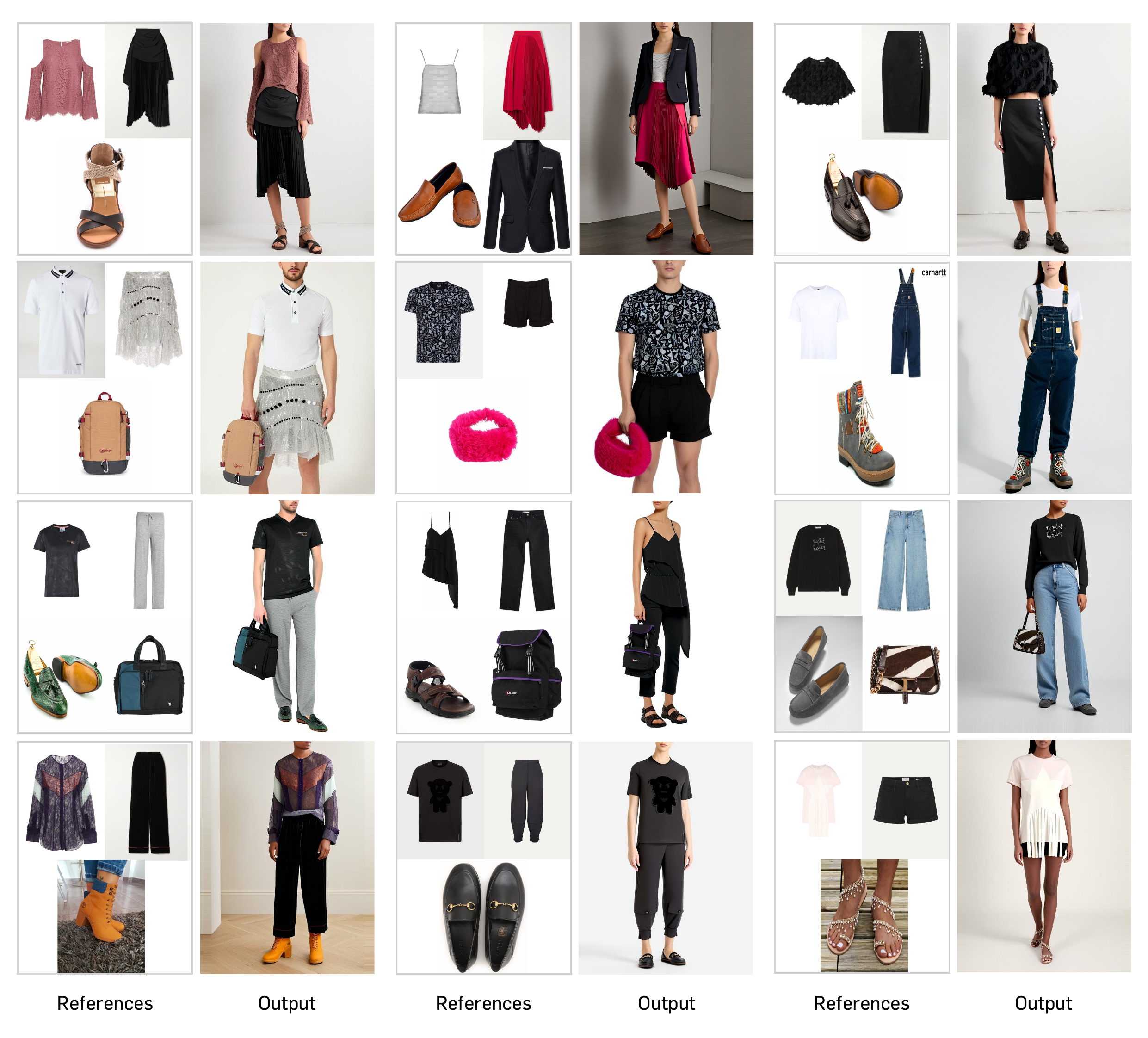}    
\captionof{figure}{\textbf{Generated human images by \sname.} \sname~can realistically dress humans with diverse categories of garments, including bags and shoes, which are not available for previous approaches. \sname~is capable of dressing complex combinations such as jackets and inner layers (First row, second column) and less common garments such as overalls (Second row, third column).
Also, \sname~can address challenging garments such as asymmetric-length garments and sandals (Third row, second column), and garments with unique details (Last row, third column).}
\label{fig:appendix_plane}
\vspace{-0.1in}
\end{figure*}

\begin{figure*}[t]
\includegraphics[width=\textwidth]{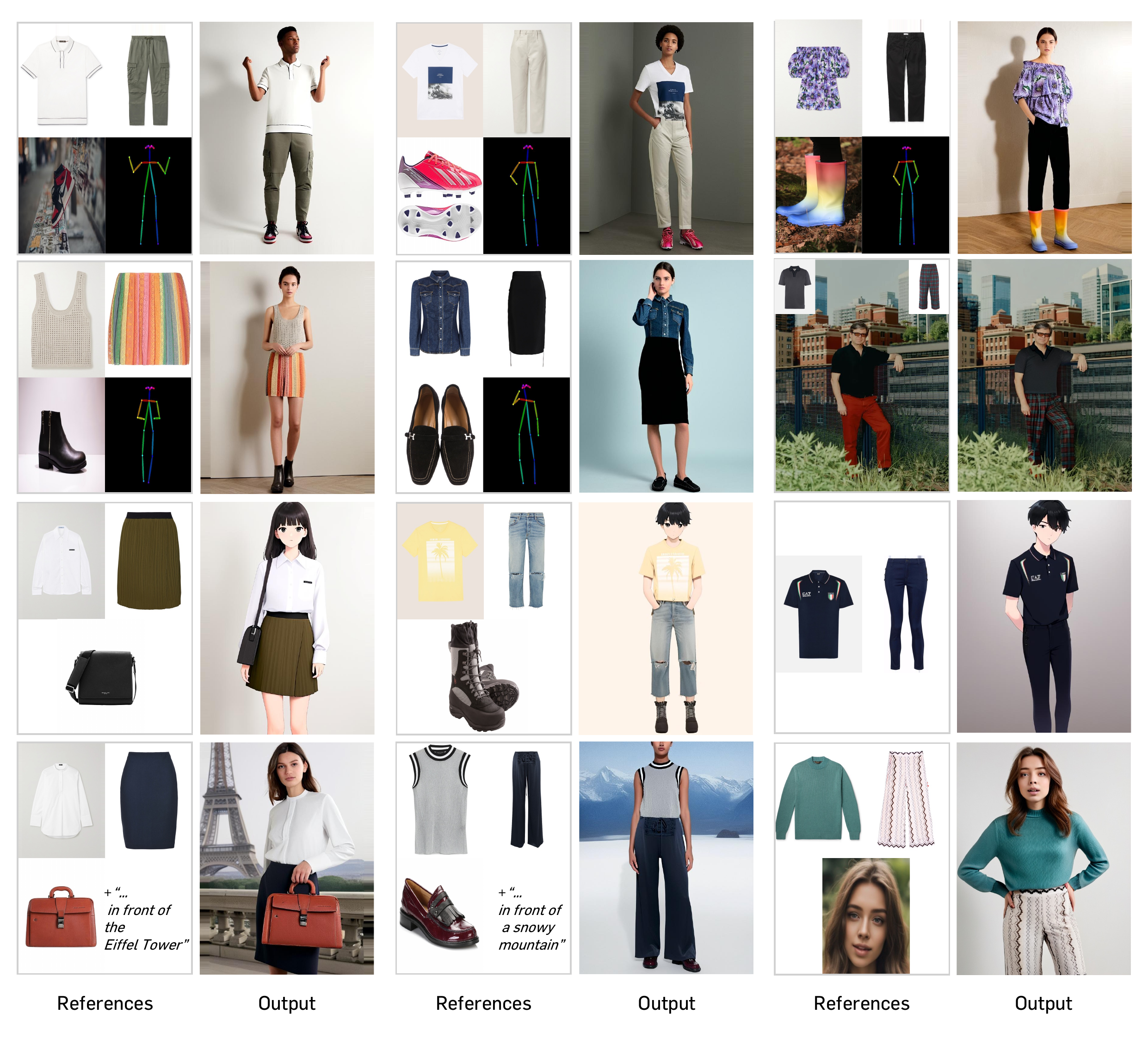} 
\captionof{figure}{\textbf{Application results by \sname.} \sname~is capable of generating human images with various conditions. By using structural conditions, it can control poses in the generated images. With text prompts, \sname~can manipulate the backgrounds of images. Additionally, it supports personalized generation through virtual try-on and face-based generations. }
\label{fig:appendix_apps}
\end{figure*}

We provide more visualizations of human images generated by \sname. We show more qualitative comparisons of \sname~with baselines in~\cref{fig:appendix_compare}. We also showcase additional human images with multiple reference garments generated by \sname~in~\cref{fig:appendix_plane} and more visualizations of application results, including controllable generation, stylization, and personalized generation in~\cref{fig:appendix_apps}.

\section{Limitations} \label{sec:appendix_limit}
While \sname~is capable of generating human images with various categories of garments, it sometimes struggles to place hats on humans naturally. This arises from the limited number of hat images in the training data. One can address this by scaling up the paired data simply using our data generation pipeline.
Also, \sname~fails to preserve tiny details such as letters, which is attributed to the limitations of the backbone model, SDXL. This can be relieved by replacing backbone to other diffusion models trained with better VAE encoders with larger number of channels.
\begin{figure}[!bh]
\includegraphics[width=\linewidth]{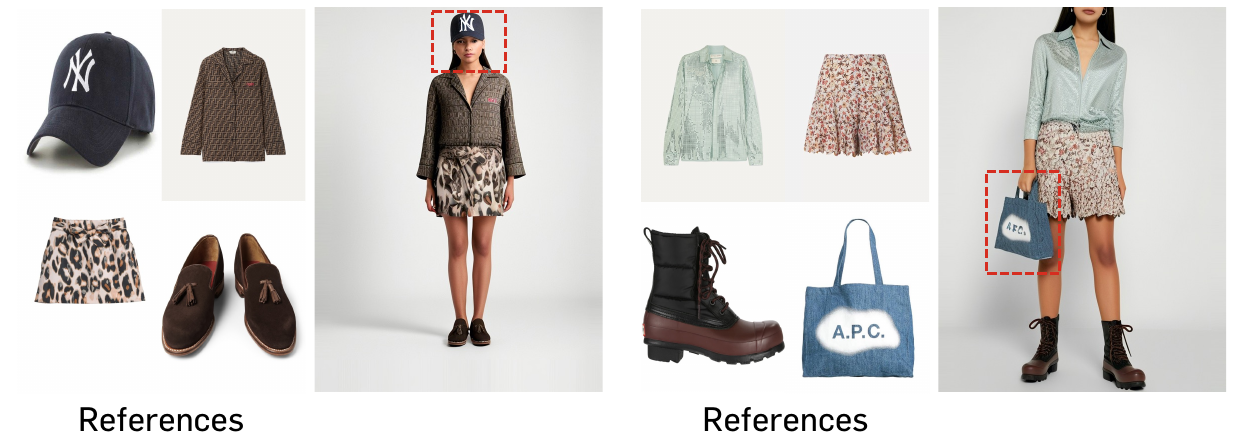}  
\captionof{figure}{\textbf{Limitations of \sname.} \sname~struggles on naturally dressing hats and preserving tiny details like letters. 
}
\vspace{-15pt}
\label{fig:hat_limitations}
\end{figure}

\clearpage

\end{document}